\documentclass[11pt,a4paper]{article}

\usepackage{polyu-vclab}

\usepackage{lipsum}                
\usepackage{amsmath}
\usepackage{booktabs}
\usepackage{multirow}
\usepackage[numbers,sort&compress]{natbib}
\usepackage{listings}
\usepackage{xcolor}

\makeatletter\def\onedot{.\@\xspace}\makeatother
\def\eg{\emph{e.g}\onedot}
\usepackage{xspace}

\definecolor{promptbg}{gray}{0.97}
\definecolor{promptrule}{gray}{0.70}
\definecolor{promptbrk}{gray}{0.45}

\lstdefinestyle{hirprompt}{
  basicstyle=\ttfamily\scriptsize,
  backgroundcolor=\color{promptbg},
  frame=leftline,
  framerule=1pt,
  rulecolor=\color{promptrule},
  framesep=6pt,
  xleftmargin=12pt,
  xrightmargin=0pt,
  breaklines=true,
  breakatwhitespace=true,
  breakautoindent=true,
  postbreak=\mbox{\textcolor{promptbrk}{$\hookrightarrow$}\space},
  columns=fullflexible,
  keepspaces=true,
  showstringspaces=false,
  aboveskip=1em,
  belowskip=1em,
  literate=
    {—}{{-{}-{}-}}3
    {–}{{-{}-}}2
    {→}{{$\rightarrow$}}1
    {ç}{{\c{c}}}1
    {“}{{``}}1
    {”}{{''}}1
    {‘}{{`}}1
    {’}{{'}}1
    {…}{{...}}3
    {©}{{\textcopyright}}1
}

\setEyebrow{PolyU VCLab\,\textbullet\,Preprint 2026}

\setReportTag{Visual Computing Lab\,\textperiodcentered\,The Hong Kong Polytechnic University}

\papertitle{HarnessIR: Harnessing Multimodal Foundation Models for Universal Real-World Image Restoration}

\paperauthors{%
  Xiangtao Kong\affilmark{1,2}\quad
  Shuaizheng Liu\affilmark{1,2}\quad
  Rongyuan Wu\affilmark{1,2}\quad
  Lingchen Sun\affilmark{1,2}\quad
  Zhengqiang Zhang\affilmark{1,2}\quad
  Jixin Zhao\affilmark{1,2}\quad
  Yuhui Wu\affilmark{1,2}\quad
  Lei Zhang\correspondmark\affilmark{1,2}%
}

\paperaffil{%
  \affilmark{1}\,The Hong Kong Polytechnic University\quad
  \affilmark{2}\,OPPO Research Institute
}

\papernotes{%
  \correspondmark\,Corresponding author (\href{mailto:cslzhang@comp.polyu.edu.hk}{cslzhang@comp.polyu.edu.hk}).%
}

\paperbadges{%
  \vclabbadgesolid{Project Page}{https://polyu-vclab.github.io/HarnessIR/}\;%
  \vclabbadgesolid{Code}{https://github.com/PolyU-VCLab/HarnessIR}\;%
  \vclabbadgesolid{Dataset}{https://huggingface.co/datasets/VCLab-PolyU/HarnessIR/tree/main}\;%
}

\begin{document}

\maketitleVCLab

\vspace{-4mm}

\begin{figure*}[h]
    \centering
    \includegraphics[width=\linewidth]{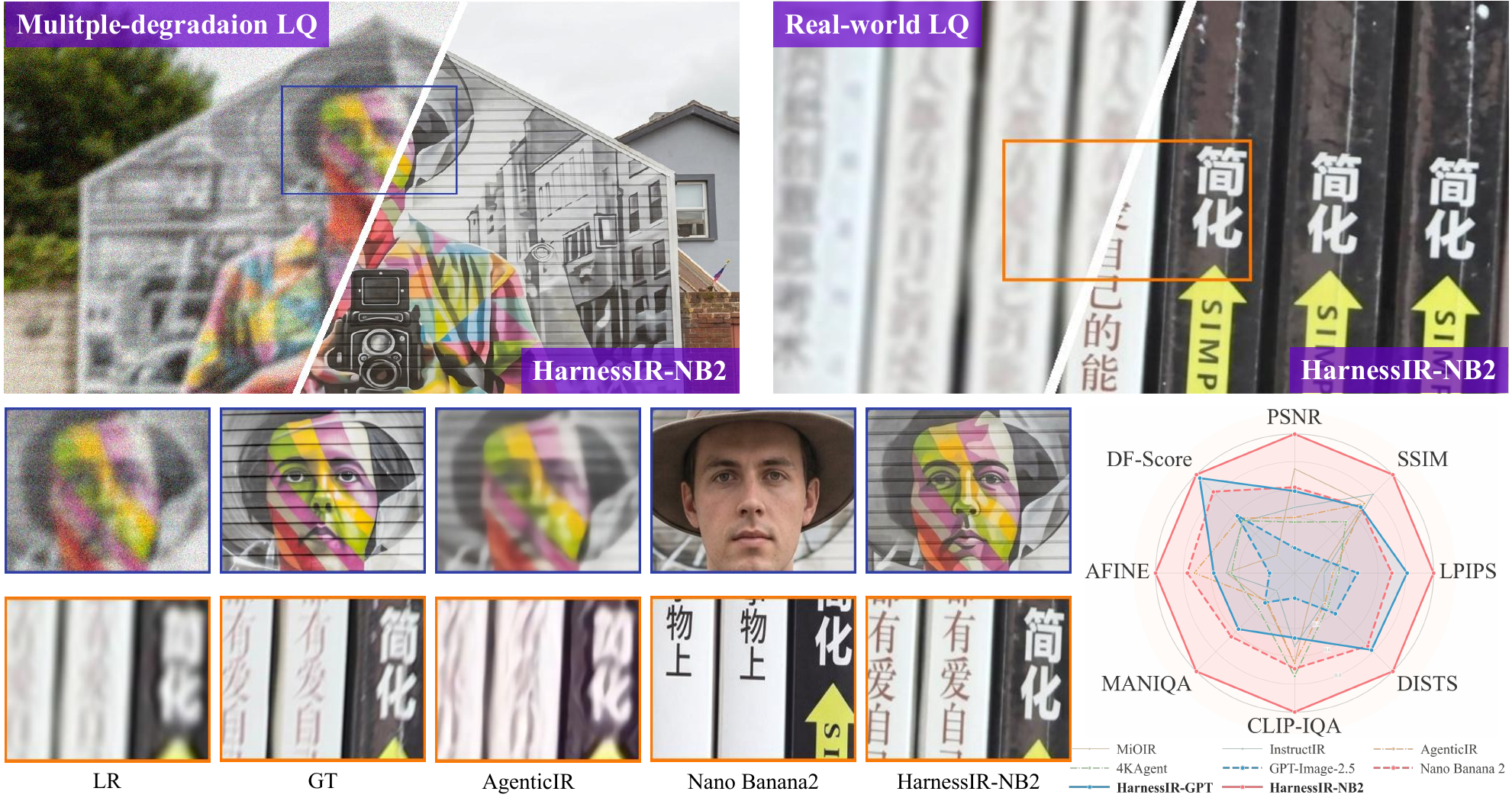}
    \vspace{-6mm}
    \caption{\textbf{Top and Bottom-left}: Visual results of HarnessIR, which can restore image details without altering their content and texts. \textbf{Bottom-right}: Radar charts comparing different metrics on real-world test sets. HarnessIR achieves much better restoration performance than Nano Banana 2, GPT-Image-2.5, previous all-in-one and agentic IR methods.}
    \label{fig:teaser}
    \vspace{-4mm}
\end{figure*}

\begin{vclabAbstract}
Real-world low-quality images suffer from complex mixed degradations, including but not limited to noise, blur, atmospheric effects, etc. Recent agentic methods usually model real-world image restoration (Real-IR) as a sequential tool calling problem over task-specific single-degradation restoration models. This paradigm, however, is fundamentally limited because complex real-world degradations cannot be cleanly undone degradation by degradation, and the tool used for task-specific models caps the capability of the agent system. In this work, we present HarnessIR, an agentic framework for Real-IR by harnessing a multimodal foundation model (MFM) as the executor. HarnessIR consists of five stages: perception and diagnosis, on-demand tool invocation, prompt composition, execution, and verification-driven refinement. Unlike prior agentic Real-IR methods that rely on tool chains assembled from task-specific models, HarnessIR feeds the restoration requirements, the perceptual diagnosis, and the evidence into an MFM that performs restoration in a single pass, followed by verification stages to determine whether the result warrants further processing. Under our harness, off-the-shelf MFMs handle restoration tasks remarkably well, achieving state-of-the-art results on the widely used MiO100 synthetic benchmark. More importantly, by exploiting the strong generalization ability of MFMs, HarnessIR delivers compelling restoration quality on challenging real-world scenes where previous agentic IR systems often struggle. Codes is available at \url{https://github.com/PolyU-VCLab/HarnessIR}. 
\end{vclabAbstract}

\section{Introduction}


Image restoration (IR) is a classical yet challenging research problem. Early IR research mainly focuses on a specific type of degradation, such as denoising~\citep{zhang2017beyond,zamir2022restormer},  deblurring~\citep{cho2021_deblur,HINet},  super-resolution~\citep{dong2014learning,chen2023hat}, dehazing~\citep{GridDehazeNet,DehazeFormer}, etc. Later, many all-in-one ~\citep{AirNet,PromptIR} or multiple-in-one~\citep{MioIR,DA-CLIP} IR methods have been proposed, aiming to train a unified model to handle different types of image degradations. However, most of these methods still assume that the input image is corrupted by one single type of degradation, and they explicitly or implicitly recognize the degradation before restoring the desired image. In practice, however, captured images in real-world are often corrupted by complex, mixed degradations that cannot be modeled by a specific degradation type.  

Inspired by the recent rapid development of agentic models in multimodal understanding and generation~\citep{gpt4,LLaVA}, agentic IR methods~\citep{agenticir,4kagent} have been proposed to address the issue of complex mixed degradations, where pretrained task-specific IR networks are exposed as \emph{restoration tools}, and {vision-language models (VLMs)} are employed {as the perceiver} to assess image content and its degradations, map the identified degradations onto specialized IR models, and {as the planner} to schedule them into an execution sequence, frequently augmented with reflection, rollback, and rescheduling. However, this formulation is subject to fundamental limitations. First, it approximates coupled mixed degradations as a sequence of specialized restoration operations, whereas many real-world degradations interact and cannot be cleanly separated or reversed in sequence. {Second, the planner converts fine-grained description of image content and degradation from the perceiver into a sequence of restoration tool calls, so much useful information is lost before restoration.} Third, task-specified IR tools generalize poorly, limiting the overall generalization ability of the whole system. Last but not least, most agentic IR methods are developed and evaluated on synthetic mixed degradations, whereas real-world degradations often interact with each other, making them difficult to simulate. Therefore, although these paradigms have achieved great progress, their generalization to real-world scenes remains limited.

Recently, generative multimodal foundation models (MFMs)~\citep{nanobanana2,gptimage25,Qwen-Image-Edit} have advanced rapidly. Having been trained for image editing, these models demonstrate great potential for IR tasks in recent works~\citep{sun2026nb2eval,yang2026realrestorer}. Basically, MFMs treat IR as an editing problem: they accept the LQ image together with an instruction to accomplish the restoration task. Unfortunately, prompting MFMs naively often yields unsatisfying results, as illustrated in Fig.~\ref{fig:teaser} and Fig.~\ref{fig:nriqa-failure}, these models often over-generate, embellishing detail and repainting texts and image contents. However, we argue that this does not mean that MFMs cannot perform complex real-world IR tasks; the key lies in how we instruct them, together with additional tools, to execute IR tasks in appropriate steps, harnessing the strong generative and generalization capabilities of MFMs for high-quality, universal real-world image restoration (Real-IR).

To this end, we present \textbf{HarnessIR}, an agentic framework for universal real-world IR by harnessing an MFM as the executor. HarnessIR is organized into five stages: perception and diagnosis, on-demand tool invocation, prompt composition, execution, and verification-driven refinement. One key distinction between our design and previous work is that HarnessIR schedules \emph{no restoration tools}, and it performs restoration by MFM calls; however, it invokes  \emph{auxiliary tools} --- OCR~\citep{ppocr}, face detection~\citep{insightface}, depth estimation~\citep{depthanythingv2}, semantic segmentation~\citep{sam3} --- which act as evidence providers for prompt composition rather than links in a restoration chain. As shown in Fig.~\ref{fig:pipeline}, the LQ image and the restoration requests are first passed to a VLM, which returns a {diagnosis} of the content and its degradations, along with a recommendation on which auxiliary tools to invoke. The system executes the plan and collects the resulting evidence, which is fed back to the VLM with the LQ input and the first-stage analysis. The VLM then composes a prompt that contains rich content and degradation information, as well as the gathered evidence rendered in language. The prompt and the LQ image are passed to the MFM executor, and the VLM then examines the output to determine whether the restoration is satisfactory and how to revise the prompt for re-execution. HarnessIR binds the MFM to what the given LQ image actually needs for restoration, eliciting the MFM's restoration capability while regulating how far it can alter the input, so that degradations can be removed without changing the content.

Using advanced MFMs such as Nano~Banana~2~\citep{nanobanana2} and GPT-Image-2.5-Sunburst~\citep{gptimage25} as executors, HarnessIR handles image restoration remarkably well. On MiO100 \citep{MioIR}, a widely used synthetic mixed-degradation benchmark, HarnessIR surpasses existing all-in-one and agentic IR methods on almost every metric. On the more challenging real-world images captured with both single and mixed degradations, where many all-in-one and agentic IR methods fail outright, HarnessIR outperforms prior methods by a large margin. Our ablation studies attribute these restoration gains to the harness rather than the executor alone. Our contributions are summarized as follows:

\begin{itemize}
\item We first identify the limitations in the previous agentic IR paradigms and propose moving the agent action primitive from restoration tool scheduling to MFM harnessing.

\item We then propose \textbf{HarnessIR}, an agentic framework that harnesses a single MFM as the restorer without using task-specific restoration models as tools.

\item Experiments show that HarnessIR reaches leading performance on the widely used MiO100 benchmark; more importantly, it restores real-world images with complex degradations significantly better than existing agentic IR methods. 
\end{itemize}

\section{Related Work and Discussion}

\vspace{2mm} \noindent\textbf{All-in-One and Agent-based Image Restoration.}
Early IR research usually addresses a specific type of degradation in isolation, such as denoising~\citep{zhang2017beyond}, deblurring~\citep{cho2021_deblur}, super-resolution~\citep{dong2014learning}, dehazing~\citep{DehazeFormer}, etc. Subsequently, many all-in-one and multi-in-one methods have been proposed to address different degradations using a unified network, such as contrastive degradation embeddings~\citep{AirNet}, degradation-aware learnable prompts~\citep{PromptIR,MioIR}, CLIP-based degradation cues~\citep{DA-CLIP, autodir}, and natural-language instructions~\citep{InstructIR}. However, most of these methods assume that the input LQ image still has a single identifiable degradation, which limits their performance in unconfined real-world scenes~\citep{realsr}, where the captured images are often corrupted by complex and mixed degradations. To handle mixed degradations, agentic IR~\citep{agenticir} takes task-specific restoration networks as tools and plans a sequence of calls over them. Existing agentic IR systems differ mainly in the design of the sequence. One line of research searches the space of tool orderings and uses intermediate results to update the plan~\citep{restoreagent,agenticir,4kagent}. Another line of work constrains the plan using degradation priors, staged decomposition, or complexity-based routing~\citep{mair,qagent,hybridagent}. The third line learns the policy directly via reinforcement learning using perceptual or no-reference quality rewards~\citep{restorer1,tiragent,jarvisir,opera}. In all these paradigms, the tools are task-specific restoration networks, which are also executors. Consequently, their generalization performance is bounded by these tools, which work poorly under complex, out-of-distribution real-world degradations.

\vspace{2mm} \noindent\textbf{Multimodal Foundation Models (MFMs).}
Generative MFMs~\citep{Qwen-Image-Edit,flux-2-2025,gptimage25} and their instruction-based editing interface allow a single model to perform an open set of pixel-level operations specified in language. Since restoration can be phrased as an editing instruction, researchers have begun to examine MFMs for IR. \citep{sun2026nb2eval} evaluated Nano Banana 2 \citep{nanobanana2} as a unified restorer. While the perceptual results are appealing, it over-synthesizes many unfaithful details. Similarly, ~\citep{yang2026realrestorer} prompted MFM with fixed instructions and found that the fidelity of the image is difficult to preserve. However, we argue that this does not mean that MFMs lack the capability for performing faithful Real-IR. The key lies in how we prompt them. With an appropriate harness to elicit and utilize their capabilities, the MFMs can significantly improve perceptual quality while preserving image content.

\vspace{2mm} \noindent\textbf{Harness: From Foundation Models to Agentic Systems.}
An MFM is not by itself an agent. What turns it into an agentic system is the \emph{harness}~\citep{sweagent}, which includes the prompts and control flow that frame the task~\citep{react}, the tools that collect evidence~\citep{toolformer}, the verification logic that decides when an output is acceptable~\citep{reflexion}, and the memory if needed~\citep{memgpt}. With an appropriate harness, the base MFM can become a markedly strong agent depending on what is wrapped around it~\citep{sweagent,openhands}. Our proposed HarnessIR is a different yet much more effective instantiation for agentic IR: it diagnoses before acting, calls auxiliary tools for evidence, explicitly constrains what should be restored, and employs a verifier to decide whether to try again based on the current results.

\section{HarnessIR}

\subsection{Problem Definition and Overview}

\begin{figure*}[t]
    \centering
    \includegraphics[width=\linewidth]{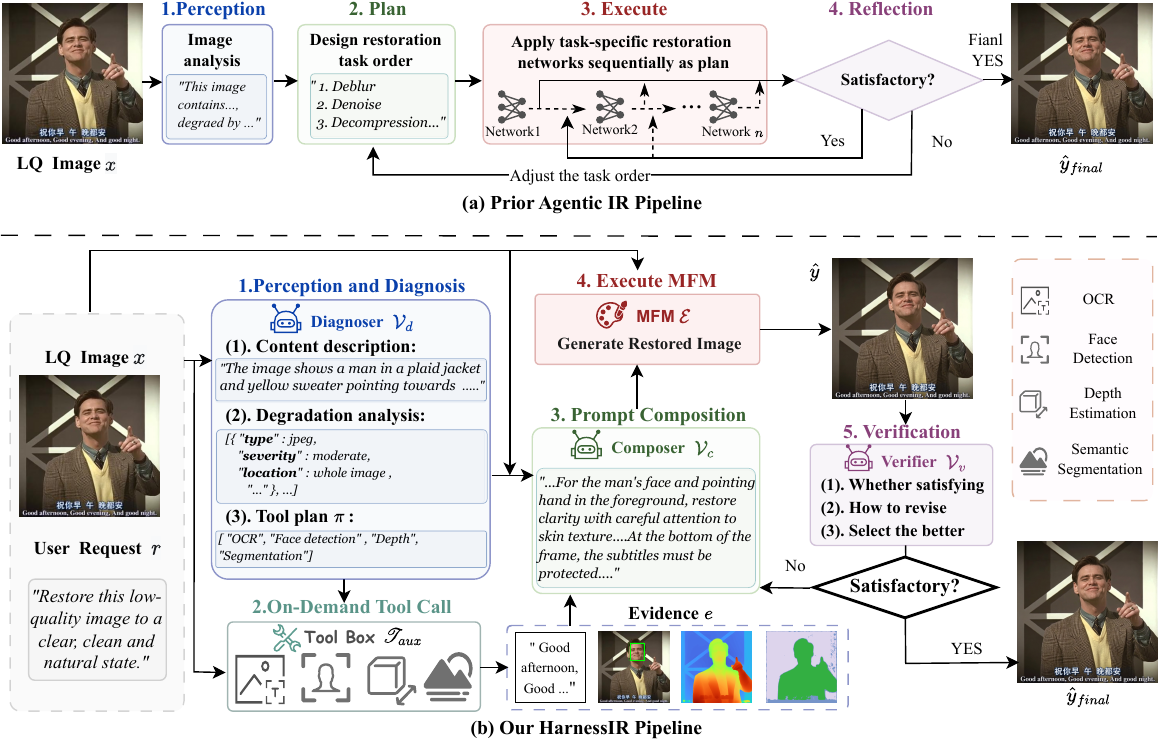}
    \caption{Comparison between (a) prior agentic IR pipelines (\textbf{top}) and (b) HarnessIR (\textbf{bottom}). {Prior methods sequentially apply task-specific restoration models planned by VLMs to produce the output. HarnessIR instead uses auxiliary tools for image-specific evidence, composes a restoration instruction for MFM, and invokes verification-driven re-execution when necessary.}}
    \label{fig:pipeline}
\end{figure*}

Given an LQ image $x$ and a user request $r$, we aim to produce a restored image $\hat{y}$. The request is a short natural-language sentence provided by the user, expressing the restoration intent ---\eg, improve the brightness of the photo (see the \textbf{Appendix~\ref{app:Request}}). As shown in Fig.~\ref{fig:pipeline}(a), prior agentic IR methods \citep{agenticir,4kagent} usually cast this as a composition problem over a library of task-specific restoration networks $\{f_t\}_{t\in\mathcal{T}}$: a perception step reads $x$ into a description, a planner turns that description into a tool order $(t_1,\dots,t_K)$, and the image is passed through the corresponding restoration networks one after another, with reflection between the steps. In contrast, our HarnessIR performs restoration with a call to a frozen MFM executor $\mathcal{E}$, driven by a prompt $p$:
\begin{equation}
\text{Prior Agentic IR:}\quad \hat{y}^{(k)}=f_{t_k}\!\big(\hat{y}^{(k-1)}\big),
\qquad\qquad
\text{HarnessIR:}\quad \hat{y}^{(k)}=\mathcal{E}\big(x,\,p^{(k)}\big),
\end{equation}
where $k=1,\dots,K$ counts rounds and $\hat{y}^{(0)}=x$. Prior agentic IR methods vary the restoration executor to process intermediate outputs sequentially. Instead, HarnessIR keeps the executor fixed and varies only the prompt. In each round, it processes the original input $x$ and therefore constitutes an alternative complete result. The verifier selects the returned candidate, $\hat{y}=\hat{y}^{(k^{\ast})}$, and requests another round only when necessary (Sec.~\ref{sec:refine}).

We use five stages to achieve our goal, as illustrated in Fig.~\ref{fig:pipeline}(b). We instantiate the diagnoser, composer, and verifier using the same frozen vision-language model (VLM) \citep{gemini37} with three role-specific briefs. The \emph{diagnoser} $\mathcal{V}_d$ produces a diagnosis $d$ of image content and degradations, together with a plan $\pi\subseteq\mathcal{T}_{\mathrm{aux}}$ specifying what auxiliary tools to consult. After the planned tools return evidence $e$, the \emph{composer} $\mathcal{V}_c$ writes $p$ from $(x,r,d,e)$; the executor $\mathcal{E}$ produces $\hat{y}$; and the \emph{verifier} $\mathcal{V}_v$ decides whether the result is accepted and which candidate is retained. Diagnosis and composition are separated because the final instruction can only be written after the diagnoser selects the tools and the requested evidence has been acquired. HarnessIR focuses on how to construct an image-specific instruction and how to verify and re-execute. In the following sections, we describe its evidence collection, prompt composition, and failure-triggered refinement in detail.


\subsection{What is Planned: Evidence to Gather, Not Restorers to Run}
\label{sec:tools}

HarnessIR plans over a different space from previous agentic IR methods \citep{agenticir,mair}, which search over orderings of restoration networks to decide how the image will be processed. Instead, HarnessIR plans what the system should perceive before writing $p$. The set of auxiliary tools --- text recognition, face detection, depth estimation, and semantic segmentation --- supplies observations that the executor would not otherwise seek on its own. For a capable MFM, the simpler the prompt it is given, the more freely it follows its own prior, which may result in hallucinations for restoration. The evidence and the instructions elicit what the executor should do to reduce the possibility for hallucination.

The tools to consult are decided by the diagnoser that produces $d$ in the same call, rather than by fixed rules, because the decision depends on the content of the image: text recognition is worth running when legible text is present, and depth estimation is worth taking when the degradation is spatially structured. For images with simple contents and degradation, the tools may not need to be called. The invoked tools return raw, image-specific observations, e.g., recognized strings, confidences, and bounding boxes for texts; bounding boxes for faces; a dense depth map; and a segmentation map with semantic labels. These outputs are supplied directly to the composer in their native structured or visual forms, and they are not passed to the executor. A complete example is provided in \textbf{Appendix~\ref{app:completecase}}.

\subsection{What is Passed to Executor: Instruction Prompt, Not Execution Order}
\label{sec:prompt}

The diagnosis is expressed in language, while the evidence from the tool may be structured or visual. The composer interprets this evidence jointly with the LQ image rather than merely transcribing it, and converts the relevant observations into a single textual instruction. The evidence is folded into the image description rather than appended as a list, each piece tied to the content it describes and the degradation it bears witness to. For example, the haze is to be cleared more aggressively in the background than in the foreground, as the depth indicates. The composer decides whether to accept the evidence or weigh the evidence against the LQ input $x$.

In addition, the composer's role-specific brief includes general restoration requirements that encode domain knowledge about each restoration operation. For example, deraining and desnowing should remove falling rain and snow without removing wet roads or snow cover in the scene, while brightness should remain unchanged when low-light enhancement is not requested. The composer jointly considers the provided information and these requirements, retains only the applicable requirements, and rewrites them into an image-specific prompt $p$. The executor receives only the LQ image $x$ and the composed prompt $p$. Unlike prior agentic IR methods, which formulate their analysis as a list of task-specific restoration tools, HarnessIR embeds image-specific information into the prompt, including what degradations to treat, what content to preserve, and where these requirements apply. The complete brief of the composer is provided in the released codes.

\subsection{When the Loop Fires: On Failure, Not by Design}
\label{sec:refine}

HarnessIR is designed to succeed in one execution and invoke another round only when the verifier judges the result unsuccessful. The output is first measured  using NR-IQA scores and inexpensive comparisons with $x$, including brightness, color statistics, and SSIM between the output and the LQ input. These measurements provide diagnostic evidence rather than fixed acceptance thresholds: they are passed to the verifier together with $x$ and the output, and the verifier makes the final decision. The verifier carries the same restoration requirements as the composer and weighs the output against them, together with the request $r$ and the measurements. The instruction to the executor is not shown to the verifier so that the result is judged against what the restoration task is supposed to do rather than against the instruction we ask the executor to follow. 

HarnessIR starts over from $x$, not the output of the last round as in previous agentic IR methods. This is because if the output of the last round is verified to be unsuccessful, the content of the image can be altered, such as incorrect text, fabricated structure, etc. Such an output is the wrong place to start from, while $x$ remains the safest input for the executor to start from. When the verifier calls for a redo, it writes down what should be corrected. The composer then writes a new instruction for the same $x$, and passes it to the executor. The verifier also compares the new restoration result with the best candidate obtained in previous rounds and retains the better one as $\hat{y}$. We allow at most three execution rounds; if no candidate is accepted by then, the best candidate selected by the verifier is returned. The complete verifier prompts are provided in the released codes.

\section{Experiments}
\label{sec:experiments}

\subsection{Experimental Setup}

\begin{figure*}[t]
    \centering
        \includegraphics[width=\linewidth]{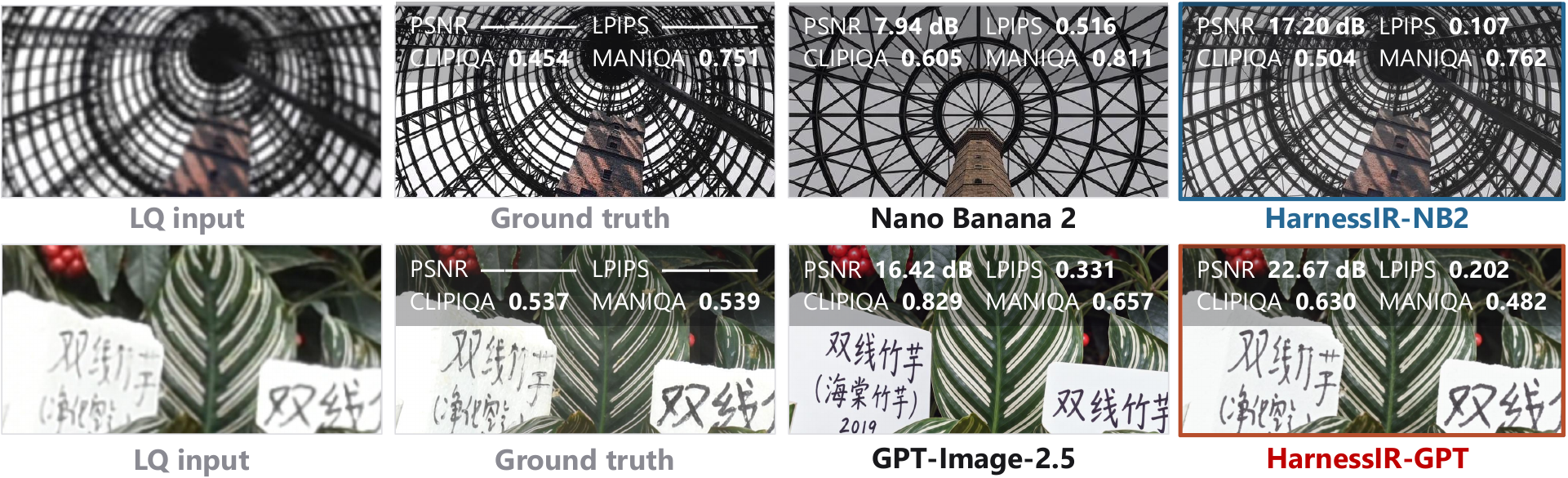}
        \caption{Direct use of MFMs may yield higher NR-IQA scores than GT, even when the image content is largely altered. NR-IQA should not be used as an independent metric for IR performance.}
        \label{fig:nriqa-failure}
\end{figure*}

\vspace{2mm} \noindent\textbf{Implementation of HarnessIR.}
HarnessIR is \textbf{inference-only}: no component is trained or fine-tuned. The diagnoser, composer, and verifier are all filled by Gemini-3.7-Flash~\citep{gemini37}, queried once per role under different briefs (the system prompts). We run HarnessIR with two MFMs: Nano~Banana~2 (NB2)~\citep{nanobanana2} and GPT-Image-2.5-Sunburst (GPT-Image)~\citep{gptimage25}, each accessed via its public interface. Four auxiliary tools are used: PP-OCRv6~\citep{ppocr} for text recognition, InsightFace Project~\citep{insightface} for face detection, Depth-Anything-V2~\citep{depthanythingv2} for depth estimation, and SAM 3~\citep{sam3} for semantic segmentation. We use a fixed request template for each degradation category to represent the usage of general users (see the \textbf{Appendix~\ref{app:Request}}).

For each executor $\mathcal{E}$, we report two variants under exactly the same LQ image and category-specific request $r$: \textbf{Baseline-$\mathcal{E}$}, which directly receives the LQ image and $r$, representing direct MFM usage; \textbf{HarnessIR-$\mathcal{E}$}, our proposed pipeline with at most three MFM execution rounds per image. The difference between the two variants lies in only the harness, rather than the user request.

\vspace{2mm} \noindent\textbf{Test Sets.}
We evaluate HarnessIR on three test sets. One is the synthetic test set MiO100~\citep{MioIR,agenticir}, which has three degradation groups (A, B, and C). Since our main concern is the model performance on real-world LQ images, following SEAR~\citep{sear}, which aggregates 100 real-world pairs, we collect two larger real-world test sets from public datasets: Real-Paired-200 with paired GT and Real-NoGT-200 without GT. They cover unconstrained mixed degradations, as well as haze, rain, snow, low light and so on. The composition and sources are detailed in the \textbf{Appendix~\ref{app:real_datasets}}, and both sets will be made publicly available. 

\vspace{2mm} \noindent\textbf{Compared Methods.}
We compare HarnessIR against six all-in-one models (AirNet~\citep{AirNet}, PromptIR~\citep{PromptIR}, MiOIR~\citep{MioIR}, DA-CLIP~\citep{DA-CLIP}, InstructIR~\citep{InstructIR}, AutoDIR~\citep{autodir}) and six agentic methods (AgenticIR~\citep{agenticir}, MAIR~\citep{mair}, 4KAgent~\citep{4kagent}, TIR-Agent~\citep{tiragent}, SEAR~\citep{sear}, OPERA~\citep{opera}). For those methods whose codes and weights are publicly available, we reproduce their results with the official implementations, and report more NR-IQA metrics beyond the ones reported in their original papers. For those methods whose codes are not released, we only quote the MiO100 results reported in their papers.

\begin{table}[t]
\centering
\caption{Group-averaged results on MiO100. The best and second-best results are marked in \textcolor{red}{red} and \textcolor{blue}{blue}, respectively. Results in which HarnessIR improves on its baseline are in \textbf{bold}. $\Delta$ means the absolute deviation from the GT score; the smaller, the better. The GT and LQ rows report the raw metric values. D-Score and F-Score are given by the VLM evaluator for degradation removal and content preservation, both on a scale of $0$--$100$; DF-Score is their per-image geometric mean.}
\label{tab:mio100_avg}
\footnotesize
\setlength{\tabcolsep}{3pt}
\resizebox{\textwidth}{!}{%
\begin{tabular}{lcccccccccccc}
\toprule
Method & \multicolumn{4}{c}{Full-reference Metrics} & \multicolumn{5}{c}{No-reference Metrics} & \multicolumn{3}{c}{VLM-based IQA} \\
\cmidrule(lr){2-5}\cmidrule(lr){6-10}\cmidrule(lr){11-13}
 & PSNR$\uparrow$ & SSIM$\uparrow$ & LPIPS$\downarrow$ & DISTS$\downarrow$ & $\Delta$MANIQA$\downarrow$ & $\Delta$CLIP-IQA$\downarrow$ & $\Delta$MUSIQ$\downarrow$ & $\Delta$TOPIQ$\downarrow$ & $\Delta$AFINE-NR$\downarrow$ & D-Score$\uparrow$ & F-Score$\uparrow$ & DF-Score$\uparrow$ \\
\midrule
AirNet & 20.30 & 0.65 & 0.46 & 0.25 & 0.22 & 0.28 & 29.00 & 0.32 & 0.27 & 25.1 & 89.7 & 29.78 \\
PromptIR & 20.66 & 0.66 & 0.45 & 0.26 & 0.21 & 0.28 & 28.98 & 0.33 & 0.26 & 26.4 & \textcolor{red}{91.7} & 30.41 \\
MiOIR & 20.76 & 0.66 & 0.43 & 0.25 & 0.22 & 0.29 & 27.29 & 0.33 & 0.26 & 28.5 & \textcolor{blue}{90.6} & 37.09 \\
DA-CLIP & 20.33 & 0.64 & 0.46 & 0.26 & 0.22 & 0.29 & 28.37 & 0.33 & 0.25 & 19.4 & 87.7 & 25.47 \\
InstructIR & 20.84 & 0.60 & 0.54 & 0.28 & 0.24 & 0.31 & 31.84 & 0.37 & 0.32 & 21.9 & \textcolor{blue}{90.6} & 29.42 \\
AutoDIR & 20.59 & 0.65 & 0.39 & 0.23 & 0.14 & 0.25 & 16.92 & 0.25 & 0.15 & 38.9 & 68.0 & 39.05 \\
\midrule
AgenticIR & 21.15 & \textcolor{blue}{0.67} & 0.35 & 0.21 & 0.14 & 0.18 & 11.73 & 0.19 & 0.07 & 49.4 & 43.3 & 42.97 \\
MAIR & 20.45 & 0.64 & 0.33 & - & 0.32 & 0.18 & 11.61 & - & - & - & - & - \\
4KAgent & 20.73 & 0.63 & 0.34 & 0.20 & 0.10 & 0.11 & 8.31 & 0.12 & 0.11 & 47.1 & 36.3 & 36.88 \\
TIR-Agent & 21.57 & 0.66 & 0.32 & - & 0.25 & 0.10 & 6.83 & - & - & - & - & - \\
SEAR & 21.51 & \textcolor{blue}{0.67} & 0.34 & - & 0.32 & 0.17 & 10.56 & - & - & - & - & - \\
OPERA & \textcolor{blue}{21.78} & \textcolor{red}{0.69} & 0.37 & - & 0.32 & 0.22 & 12.67 & - & - & - & - & - \\
\midrule
GPT-Image-2.5 & 17.95 & 0.54 & 0.36 & 0.20 & 0.02 & 0.10 & 4.64 & 0.06 & \textcolor{blue}{0.01} & 48.8 & 20.9 & 30.38 \\
HarnessIR-GPT & \textbf{19.70} & \textbf{0.58} & \textbf{0.30} & \textbf{0.17} & \textcolor{blue}{\textbf{0.01}} & \textcolor{blue}{\textbf{0.07}} & \textbf{2.81} & \textcolor{red}{\textbf{0.02}} & \textcolor{blue}{\textbf{0.01}} & \textbf{67.5} & \textbf{37.7} & \textbf{48.44} \\
\midrule
Nano Banana 2 & 20.81 & 0.63 & \textcolor{blue}{0.24} & \textcolor{blue}{0.13} & \textcolor{red}{0.00} & \textcolor{red}{0.03} & \textcolor{blue}{0.70} & \textcolor{blue}{0.04} & \textcolor{red}{0.00} & \textcolor{red}{76.7} & 44.9 & \textcolor{blue}{55.87} \\
HarnessIR-NB2 & \textcolor{red}{\textbf{22.21}} & \textbf{0.65} & \textcolor{red}{\textbf{0.21}} & \textcolor{red}{\textbf{0.12}} & \textcolor{red}{0.00} & \textcolor{red}{0.03} & \textcolor{red}{\textbf{0.25}} & \textcolor{blue}{0.04} & \textcolor{red}{0.00} & \textcolor{blue}{75.2} & \textbf{51.7} & \textcolor{red}{\textbf{59.94}} \\
\midrule
GT & - & - & - & - & 0.64 & 0.65 & 68.79 & 0.64 & -0.92 & - & - & - \\
LQ & 19.86 & 0.57 & 0.58 & 0.31 & 0.23 & 0.28 & 32.77 & 0.37 & 0.31 & - & - & - \\
\bottomrule
\end{tabular}}
\end{table}

\subsection{Evaluation Protocol and Metrics}
Following prior works~\citep{agenticir,yang2026realrestorer,sear}, we measure fidelity with PSNR and SSIM~\citep{ssim} in the Y channel, perceptual quality with LPIPS~\citep{lpips} and DISTS~\citep{dists}, and image quality with the no-reference image quality assessment (NR-IQA) metrics MANIQA~\citep{maniqa}, CLIP-IQA~\citep{clipiqa}, MUSIQ~\citep{musiq}, TOPIQ~\citep{chen2024topiq}, and A-FINE~\citep{afine}. However, it should be noted that a higher NR-IQA score does not, by itself, indicate better restoration performance because their purpose is to evaluate image quality without considering whether the restored content is correct or not. As a result, hallucination can be rewarded, even when the content of the restored image deviates significantly from the input LQ image. Fig.~\ref{fig:nriqa-failure} shows some examples. We see that the restored images significantly outscore their GTs on NR-IQA metrics, while the image content and text have been largely repainted. Such cases are rare for earlier discriminative restoration networks, but can be common for diffusion-based generative models. This is because the generative methods are more prone to altering image content and NR-IQA metrics only measure whether the given image is of good quality, no matter what content it has.

To solve this issue of unreasonable use of NR-IQA metrics in Real-IR method evaluation, we make two adjustments. First, when a GT exists, we report the absolute distance, denoted by $\Delta$, between the NR-IQA scores of the restored image and the GT image. A value of $\Delta$ closer to zero means better quality since the GT is the target. Second, regardless of whether a GT exists, following recent IR and IQA works~\citep{PixRestore,VQR1}, we use an independent VLM evaluator of the HarnessIR system to assess restoration performance directly. Given the LQ input and the restored output, it reports D-Score for degradation removal and F-Score for content preservation, both on a scale of $0$--$100$. For each image, we compute their geometric mean, DF-Score $=\sqrt{D F}$, which is high when both degradation removal and fidelity are good. The evaluator prompt and protocol, ablations, and comparisons with human judgments are provided in the \textbf{Appendix~\ref{app:dfscore}}.

\begin{table}[t]
\centering
\caption{Results on Real-Paired-200, which contains 200 real-world LQ images with paired GT from public real-world datasets. Marking and metrics follow Tab.~\ref{tab:mio100_avg}.}
\label{tab:real_paired_200}
\footnotesize
\setlength{\tabcolsep}{4pt}
\resizebox{\textwidth}{!}{%
\begin{tabular}{lcccccccccccc}
\toprule
Method & \multicolumn{4}{c}{Full-reference Metrics} & \multicolumn{5}{c}{No-reference Metrics} & \multicolumn{3}{c}{VLM-based IQA} \\
\cmidrule(lr){2-5}\cmidrule(lr){6-10}\cmidrule(lr){11-13}
 & PSNR$\uparrow$ & SSIM$\uparrow$ & LPIPS$\downarrow$ & DISTS$\downarrow$ & $\Delta$MANIQA$\downarrow$ & $\Delta$CLIP-IQA$\downarrow$ & $\Delta$MUSIQ$\downarrow$ & $\Delta$TOPIQ$\downarrow$ & $\Delta$AFINE-NR$\downarrow$ & D-Score$\uparrow$ & F-Score$\uparrow$ & DF-Score$\uparrow$ \\
\midrule
AirNet & 22.40 & 0.724 & 0.440 & 0.254 & 0.108 & 0.107 & 13.27 & 0.109 & 0.133 & 16.8 & 85.7 & 25.2 \\
PromptIR & 24.08 & 0.743 & 0.416 & 0.241 & 0.098 & 0.099 & 12.78 & 0.105 & 0.114 & 16.7 & \textcolor{blue}{91.3} & 25.3 \\
MiOIR & \textcolor{blue}{24.87} & 0.735 & 0.418 & 0.251 & 0.096 & 0.086 & 12.20 & 0.104 & 0.124 & 10.5 & \textcolor{red}{93.0} & 18.9 \\
DA-CLIP & 23.87 & 0.739 & 0.406 & 0.243 & 0.099 & 0.106 & 11.83 & 0.093 & 0.111 & 17.8 & 88.2 & 26.7 \\
InstructIR & 22.92 & \textcolor{blue}{0.747} & 0.411 & 0.236 & 0.108 & 0.116 & 12.34 & 0.104 & 0.124 & 25.4 & 90.2 & 36.9 \\
AutoDIR & 23.80 & 0.735 & 0.372 & 0.233 & 0.079 & 0.093 & \textcolor{blue}{4.19} & 0.052 & \textcolor{blue}{0.046} & 29.0 & 74.9 & 32.5 \\
\midrule
AgenticIR & 22.41 & 0.734 & 0.398 & 0.235 & 0.097 & 0.087 & 7.41 & 0.072 & 0.083 & 33.7 & 68.4 & 38.8 \\
4KAgent & 22.18 & 0.708 & 0.389 & 0.231 & 0.088 & \textcolor{blue}{0.064} & \textcolor{red}{2.89} & \textcolor{blue}{0.030} & 0.131 & 35.2 & 56.1 & 37.4 \\
\midrule
GPT-Image-2.5 & 20.87 & 0.660 & 0.361 & 0.222 & 0.095 & 0.201 & 15.78 & 0.212 & 0.180 & 69.0 & 27.5 & 40.1 \\
HarnessIR-GPT & \textbf{23.74} & \textbf{0.729} & \textcolor{blue}{\textbf{0.286}} & \textcolor{blue}{\textbf{0.176}} & \textbf{0.066} & \textbf{0.131} & \textbf{12.32} & \textbf{0.128} & \textbf{0.108} & \textcolor{red}{\textbf{82.7}} & \textbf{50.4} & \textcolor{blue}{\textbf{60.2}} \\
\midrule
Nano Banana 2 & 23.94 & 0.729 & 0.309 & 0.181 & \textcolor{blue}{0.058} & 0.077 & 11.91 & 0.118 & 0.074 & 67.9 & 47.8 & 52.9 \\
HarnessIR-NB2 & \textcolor{red}{\textbf{26.63}} & \textcolor{red}{\textbf{0.775}} & \textcolor{red}{\textbf{0.246}} & \textcolor{red}{\textbf{0.149}} & \textcolor{red}{\textbf{0.020}} & \textcolor{red}{\textbf{0.001}} & \textbf{5.52} & \textcolor{red}{\textbf{0.028}} & \textcolor{red}{\textbf{0.033}} & \textcolor{blue}{\textbf{75.2}} & \textbf{63.6} & \textcolor{red}{\textbf{62.0}} \\
\midrule
GT & - & - & - & - & 0.560 & 0.451 & 54.36 & 0.417 & -0.901 & - & - & - \\
LQ & 25.27 & 0.736 & 0.436 & 0.258 & 0.098 & 0.092 & 13.08 & 0.107 & 0.115 & - & - & - \\
\bottomrule
\end{tabular}}
\end{table}

\begin{table}[t]
\centering
\footnotesize
\setlength{\tabcolsep}{4pt}
\caption{Results on Real-NoGT-200. Marking and metrics follow Tab.~\ref{tab:mio100_avg}. Without GT, NR-IQA alone cannot correctly reflect the restoration performance, so these columns are \textcolor{gray}{grayed out}.}

\resizebox{\linewidth}{!}{%
\begin{tabular}{lcccccccc}
\toprule
Method & \multicolumn{5}{c}{No-reference Metrics} & \multicolumn{3}{c}{VLM-based IQA} \\
\cmidrule(lr){2-6}\cmidrule(lr){7-9}
 & MANIQA$\uparrow$ & CLIP-IQA$\uparrow$ & MUSIQ$\uparrow$ & TOPIQ$\uparrow$ & AFINE-NR$\downarrow$ & D-Score$\uparrow$ & F-Score$\uparrow$ & DF-Score$\uparrow$ \\
\midrule
AirNet & \textcolor{gray}{0.525} & \textcolor{gray}{0.406} & \textcolor{gray}{42.94} & \textcolor{gray}{0.307} & \textcolor{gray}{-0.689} & 13.4 & 86.5 & 21.6 \\
PromptIR & \textcolor{gray}{0.531} & \textcolor{gray}{0.413} & \textcolor{gray}{43.17} & \textcolor{gray}{0.310} & \textcolor{gray}{-0.703} & 11.2 & \textcolor{red}{90.4} & 21.0 \\
MiOIR & \textcolor{gray}{0.537} & \textcolor{gray}{0.399} & \textcolor{gray}{44.44} & \textcolor{gray}{0.321} & \textcolor{gray}{-0.695} & 11.4 & \textcolor{blue}{90.2} & 22.5 \\
DA-CLIP & \textcolor{gray}{0.529} & \textcolor{gray}{0.395} & \textcolor{gray}{43.50} & \textcolor{gray}{0.311} & \textcolor{gray}{-0.700} & 12.6 & 88.3 & 20.3 \\
InstructIR & \textcolor{gray}{0.521} & \textcolor{gray}{0.383} & \textcolor{gray}{43.39} & \textcolor{gray}{0.311} & \textcolor{gray}{-0.689} & 21.4 & 86.0 & 32.0 \\
AutoDIR & \textcolor{gray}{0.538} & \textcolor{gray}{0.399} & \textcolor{gray}{47.85} & \textcolor{gray}{0.337} & \textcolor{gray}{-0.732} & 17.9 & 83.7 & 27.0 \\
\midrule
AgenticIR & \textcolor{gray}{0.525} & \textcolor{gray}{0.374} & \textcolor{gray}{45.38} & \textcolor{gray}{0.325} & \textcolor{gray}{-0.697} & 19.1 & 78.1 & 26.9 \\
4KAgent & \textcolor{gray}{0.516} & \textcolor{gray}{0.396} & \textcolor{gray}{48.29} & \textcolor{gray}{0.343} & \textcolor{gray}{-0.677} & 18.9 & 70.4 & 25.5 \\
\midrule
GPT-Image-2.5 & \textcolor{gray}{0.676} & \textcolor{gray}{0.693} & \textcolor{gray}{72.92} & \textcolor{gray}{0.658} & \textcolor{gray}{-1.059} & 59.3 & 16.9 & 29.6 \\
HarnessIR-GPT & \textcolor{gray}{0.660} & \textcolor{gray}{0.665} & \textcolor{gray}{70.28} & \textcolor{gray}{0.610} & \textcolor{gray}{-1.000} & \textcolor{red}{\textbf{76.4}} & \textbf{36.0} & \textcolor{blue}{\textbf{49.3}} \\
\midrule
Nano Banana 2 & \textcolor{gray}{0.623} & \textcolor{gray}{0.534} & \textcolor{gray}{66.18} & \textcolor{gray}{0.531} & \textcolor{gray}{-0.933} & 65.1 & 34.4 & 47.3 \\
HarnessIR-NB2 & \textcolor{gray}{0.605} & \textcolor{gray}{0.484} & \textcolor{gray}{62.08} & \textcolor{gray}{0.483} & \textcolor{gray}{-0.904} & \textcolor{blue}{\textbf{71.3}} & \textbf{64.1} & \textcolor{red}{\textbf{62.9}} \\
\bottomrule
\end{tabular}}
\label{tab:real_nogt_200}
\end{table}

\subsection{Main Results}

\vspace{2mm} \noindent\textbf{Results on Synthetic MiO100 Test Sets.} Tab.~\ref{tab:mio100_avg} summarizes the group-averaged results on MiO100 (detailed results of Group A, B and C are in \textbf{Appendix~\ref{app:mio100}}). Compared with the direct-use baselines (Nano Banana 2 and GPT-Image-2.5), HarnessIR-NB2 and HarnessIR-GPT gain $1.40$ and $1.75$\,dB in PSNR, improve LPIPS by $0.03$ and $0.06$, and DF-Score by $4.1$ and $18.1$ points, respectively. These gains mainly reflect better content preservation. HarnessIR also substantially improves GPT-Image-2.5's D-Score. For NB2, the D-Score changes only slightly because MiO100 is constructed by applying controlled synthetic degradations to curated GT images. NB2 can already remove these degradations effectively.

Against prior methods, HarnessIR-NB2 achieves the best results on most metrics except SSIM and F-Score. HarnessIR-GPT achieves the best result on $\Delta$TOPIQ. Several all-in-one models achieve relatively high F-Scores but D-Scores below $30$, suggesting that their fidelity comes partly from making only limited restoration changes. Although HarnessIR-GPT improves substantially over the direct use of GPT-Image-2.5, it is weak in PSNR and SSIM. This reflects the executor's capabilities rather than a limitation of the harness: the harness helps to unleash the executor's restoration potential but cannot raise its upper ceiling. GPT-Image-2.5 has a strong generative prior but tends to restore more aggressively, which makes its outputs less pixel-aligned with the GT.

\vspace{2mm} \noindent\textbf{Results on Real-Paired-200 and Real-NoGT-200 Test Sets.}
The advantage of HarnessIR is even more pronounced on real-world data. On Real-Paired-200 (Tab.~\ref{tab:real_paired_200}), both HarnessIR variants substantially improve their direct-use baselines in fidelity and restoration quality. HarnessIR-NB2 achieves a $2.69$\,dB PSNR gain and reaches a DF-Score of $62.0$, while HarnessIR-GPT improves PSNR by $2.87$\,dB and raises its DF-Score from $40.1$ to $60.2$. HarnessIR also outperforms all prior methods by a clear margin in DF-Score, with both variants exceeding the previous best of $38.8$. HarnessIR-NB2 also achieves the smallest deviation from the GT on four of the five NR-IQA metrics. In particular, HarnessIR-NB2 achieves the highest PSNR, surpassing the best prior result by $1.76$\,dB. This is notable because the LQ input already reaches $25.27$\,dB; the high PSNR but low D-Scores of several all-in-one methods therefore likely reflect limited restoration, whereas HarnessIR-NB2 improves upon the input while maintaining strong degradation removal.

The results on Real-NoGT-200 (Tab.~\ref{tab:real_nogt_200}) lead to the same conclusion. HarnessIR substantially improves both D-Score and F-Score for the two executors, yielding DF-Scores of $62.9$ for NB2 and $49.3$ for GPT-Image-2.5. These scores are well above the best prior result of $32.0$. The lower NR-IQA scores of HarnessIR relative to the direct use of MFM should therefore not be interpreted as a decline in performance. As discussed in Fig.~\ref{fig:nriqa-failure}, NR-IQA can reward hallucinated details even when the image content has been altered. We consequently use DF-Score for the no-GT setting, as it evaluates degradation removal and content preservation jointly.

\begin{figure*}[t]
    \centering
        \includegraphics[width=\linewidth]{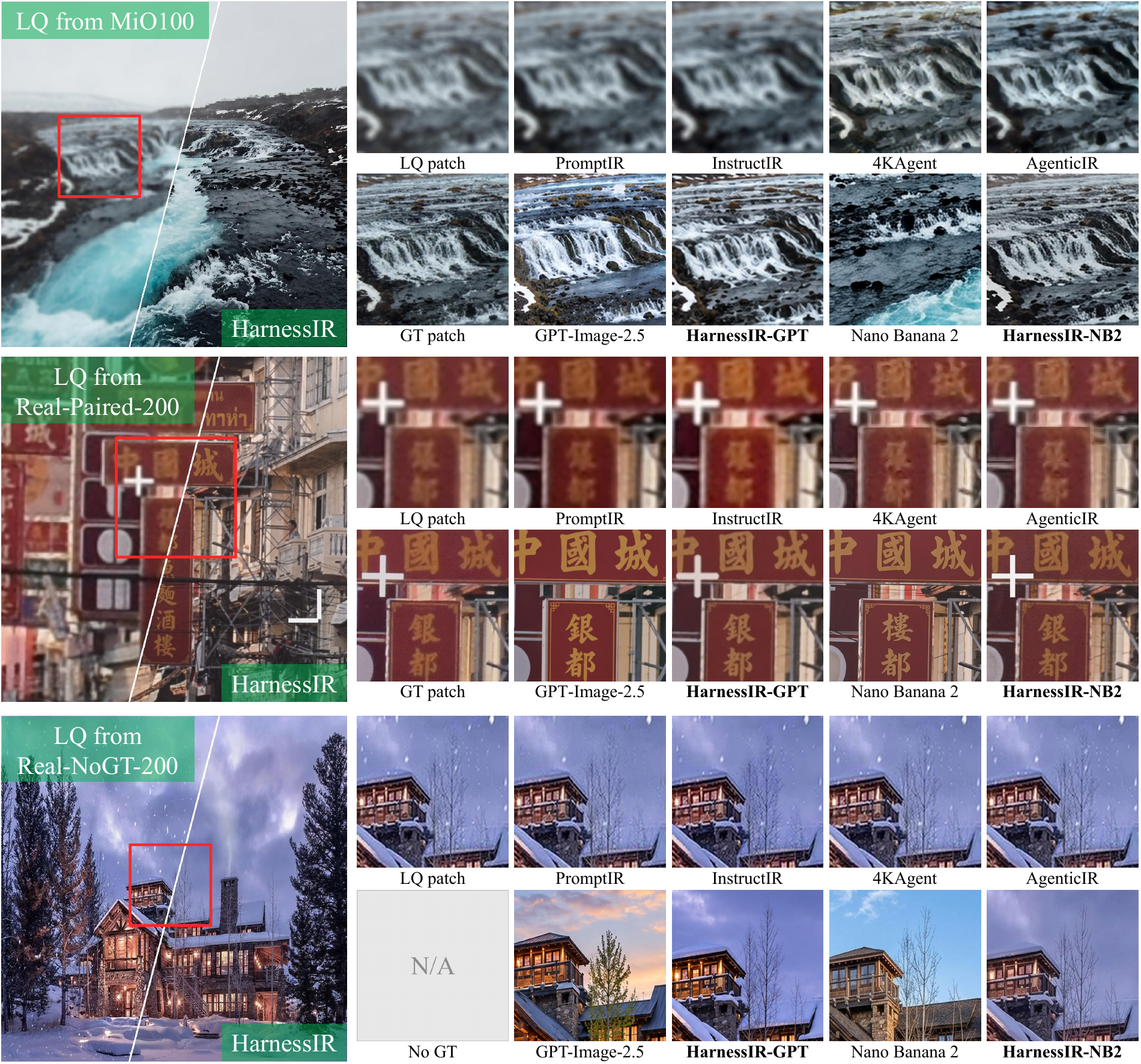}
        \caption{Visual comparisons on three test sets. HarnessIR achieves better fidelity and visual quality.}
        \label{fig:visual}
\end{figure*}

\vspace{2mm} \noindent\textbf{Visual Results.}
Fig.~\ref{fig:visual} shows qualitative comparisons across the three test sets. The all-in-one methods make little visible improvement over the LQ inputs in these examples, with only PromptIR removing some of the snow in the third image. Previous agentic IR methods make the MiO100 image somewhat clearer, but produce only limited improvement on the two real-world images. In contrast, the MFM-based methods remove degradations more effectively. However, direct use of NB2 and GPT-Image-2.5 often alters image content. In the first example, HarnessIR better preserves the original color tone and scene content. In the second, both direct MFM baselines mistakenly remove the white cross already in the image, and NB2 also changes the text. In the third, both baselines turn the nighttime scene into daytime and remove the snow cover; GPT-Image-2.5 additionally makes the trees appear to have sprouted leaves. HarnessIR avoids these unwanted changes while still removing degradations, producing visually cleaner results that remain faithful to the original LQ images.

\subsection{Ablation Studies}
\label{app:ablation}

\subsubsection{Ablation of Executor Inputs}
\label{app:ablation_prompt}

We first explore how information is prepared for and supplied to the MFM executor on Real-Paired-200. The settings vary depending on the user request, image-specific diagnosis, auxiliary-tool evidence, and whether that evidence is supplied as visual outputs or expressed in text. We use \emph{executor inputs} to refer to this information; the LQ image is provided in every setting. To isolate their effects, all variants use a single execution round, without verification and re-execution. Tab.~\ref{tab:instruction_ablation} compares five settings. In the baseline, the executor receives only the LQ image $x$ and the user's request $r$. \emph{Detailed User Request} replaces the short request with a more explicit paragraph that a user could plausibly provide, but adds no image-specific analysis. \emph{Single-VLM Instruction} uses one VLM to diagnose the image and compose the instruction without using any auxiliary tools. The final two settings follow HarnessIR's use of separate VLM stages and auxiliary tools. In \emph{Visual Evidence}, the executor also receives the tools' original outputs, including the depth map and semantic segmentation map. In \emph{Text Evidence}, the relevant evidence is composed into language by the composer, which matches the executor-input setting used by HarnessIR.

\begin{table*}[t]
\centering
\footnotesize
\setlength{\tabcolsep}{3pt}
\caption{Ablation of executor inputs on Real-Paired-200. All variants use one execution round, without re-execution to isolate the effect of how information is supplied to the executor. The \textbf{bold} row uses the executor-input setting adopted by HarnessIR. Metrics follow Tab.~\ref{tab:mio100_avg}; the best and second-best results in each column are marked in \textcolor{red}{red} and \textcolor{blue}{blue}.}
\label{tab:instruction_ablation}
\resizebox{\textwidth}{!}{%
\begin{tabular}{lcccccccccccc}
\toprule
Executor-input configuration & \multicolumn{4}{c}{Full-reference Metrics} & \multicolumn{5}{c}{No-reference Metrics} & \multicolumn{3}{c}{VLM-based IQA} \\
\cmidrule(lr){2-5}\cmidrule(lr){6-10}\cmidrule(lr){11-13}
 & PSNR$\uparrow$ & SSIM$\uparrow$ & LPIPS$\downarrow$ & DISTS$\downarrow$ & $\Delta$MANIQA$\downarrow$ & $\Delta$CLIP-IQA$\downarrow$ & $\Delta$MUSIQ$\downarrow$ & $\Delta$TOPIQ$\downarrow$ & $\Delta$AFINE-NR$\downarrow$ & D-Score$\uparrow$ & F-Score$\uparrow$ & DF-Score$\uparrow$ \\
\midrule
Direct MFM (Baseline) & 23.94 & 0.729 & 0.309 & 0.181 & 0.058 & 0.077 & 11.91 & 0.118 & 0.074 & 67.9 & 47.8 & 52.9 \\
Detailed User Request & 24.75 & 0.733 & 0.298 & 0.177 & 0.054 & 0.069 & 10.88 & 0.105 & 0.065 & 71.4 & 50.9 & 54.5 \\
\midrule
Single-VLM Instruction & 25.49 & \textcolor{blue}{0.761} & \textcolor{blue}{0.263} & \textcolor{blue}{0.161} & 0.042 & 0.044 & 9.30 & 0.080 & 0.057 & 72.5 & 54.2 & 56.6 \\
Two VLMs + Tools (Visual Evidence) & \textcolor{blue}{25.62} & 0.758 & 0.267 & \textcolor{blue}{0.161} & \textcolor{red}{0.026} & \textcolor{blue}{0.032} & \textcolor{red}{7.39} & \textcolor{red}{0.057} & \textcolor{blue}{0.049} & \textcolor{red}{74.4} & \textcolor{blue}{55.3} & \textcolor{blue}{57.8} \\
\textbf{Two VLMs + Tools (Text Evidence)} & \textcolor{red}{26.01} & \textcolor{red}{0.763} & \textcolor{red}{0.261} & \textcolor{red}{0.158} & \textcolor{blue}{0.032} & \textcolor{red}{0.024} & \textcolor{blue}{7.72} & \textcolor{blue}{0.062} & \textcolor{red}{0.045} & \textcolor{blue}{73.0} & \textcolor{red}{58.5} & \textcolor{red}{59.6} \\
\midrule
GT & - & - & - & - & 0.560 & 0.451 & 54.36 & 0.417 & -0.901 & - & - & - \\
\bottomrule
\end{tabular}}
\end{table*}

The results show a steady improvement as progressively more information is supplied to the executor. Expanding the user request alone raises PSNR from $23.94$ to $24.75$\,dB and DF-Score from $52.9$ to $54.5$. Image-specific analysis by a single VLM further increases the DF-Score to $56.6$. Adding auxiliary tools yields further gains, particularly in degradation removal: the visual-evidence setting achieves a D-Score of $74.4$ and a DF-Score of $57.8$. However, giving the executor visual maps directly can introduce artifacts, such as semantic-label colors appearing in the restored image. Expressing the relevant evidence in text avoids this risk and suits the executor's language-based interface. This setting achieves the strongest one-round result, with a PSNR of $26.01$\,dB and a DF-Score of $59.6$. Overall, the results suggest that richer requests, image-specific analysis, auxiliary evidence, and evidence-grounded instruction writing each contribute to better restoration and content preservation.

These gains require only a modest increase in inference cost. As shown in Tab.~\ref{tab:cost}, adding one VLM call raises the cost from $\$0.070$ for direct MFM usage to $\$0.077$. Using two VLM calls and auxiliary tools costs $\$0.085$. Thus, the improvements are achieved without a prohibitively expensive control loop, offering a favorable trade-off between restoration quality and cost.

\subsubsection{Redo vs. In-place Repair During Re-execution}
\label{app:ablation_redo}

We compare two refinement strategies with the single-pass version in Tab.~\ref{tab:redo_ablation}. \emph{HarnessIR (Single-pass)} performs one execution without re-execution and serves as the reference setting. \emph{HarnessIR (In-place refinement)} feeds the output of the previous execution back to the executor for further correction, following the sequential refinement strategy of previous agent IR works~\citep{agenticir,mair}. In contrast, \emph{HarnessIR (Redo from LQ)} returns to the original LQ image when refinement is triggered, composes a revised instruction, and generates a new candidate from scratch, which corresponds to the final HarnessIR design described in Sec.~\ref{sec:refine}.

\begin{table*}[t]
\centering
\footnotesize
\setlength{\tabcolsep}{3pt}
\caption{Ablation of refinement strategies on Real-Paired-200. We compare three settings: \emph{HarnessIR (Single-pass)}, which performs a single MFM execution without re-execution; \emph{HarnessIR (In-place refinement)}, which applies a revised instruction to the output of the previous round; and \emph{HarnessIR (Redo from LQ)} which applies a revised instruction to original LQ input. Metrics follow Tab.~\ref{tab:mio100_avg}, the best and second-best results are marked in \textcolor{red}{red} and \textcolor{blue}{blue}.}
\label{tab:redo_ablation}
\resizebox{\textwidth}{!}{%
\begin{tabular}{lcccccccccccc}
\toprule
Refinement strategy & \multicolumn{4}{c}{Full-reference Metrics} & \multicolumn{5}{c}{No-reference Metrics} & \multicolumn{3}{c}{VLM-based IQA} \\
\cmidrule(lr){2-5}\cmidrule(lr){6-10}\cmidrule(lr){11-13}
 & PSNR$\uparrow$ & SSIM$\uparrow$ & LPIPS$\downarrow$ & DISTS$\downarrow$ & $\Delta$MANIQA$\downarrow$ & $\Delta$CLIP-IQA$\downarrow$ & $\Delta$MUSIQ$\downarrow$ & $\Delta$TOPIQ$\downarrow$ & $\Delta$AFINE-NR$\downarrow$ & D-Score$\uparrow$ & F-Score$\uparrow$ & DF-Score$\uparrow$ \\
\midrule
HarnessIR (Single-pass) & \textcolor{blue}{26.01} & \textcolor{blue}{0.763} & \textcolor{blue}{0.261} & \textcolor{blue}{0.158} & \textcolor{blue}{0.032} & 0.024 & 7.72 & 0.062 & 0.045 & \textcolor{blue}{73.0} & \textcolor{blue}{58.5} & \textcolor{blue}{59.6} \\
HarnessIR (In-place refinement) & 25.64 & 0.760 & 0.265 & 0.159 & \textcolor{red}{0.020} & \textcolor{blue}{0.016} & \textcolor{blue}{6.79} & \textcolor{blue}{0.052} & \textcolor{red}{0.032} & 71.8 & 55.6 & 57.2 \\
HarnessIR (Redo from LQ) & \textcolor{red}{26.63} & \textcolor{red}{0.775} & \textcolor{red}{0.246} & \textcolor{red}{0.149} & \textcolor{red}{0.020} & \textcolor{red}{0.001} & \textcolor{red}{5.52} & \textcolor{red}{0.028} & \textcolor{blue}{0.033} & \textcolor{red}{75.2} & \textcolor{red}{63.6} & \textcolor{red}{62.0} \\
\midrule
GT  & - & - & - & - & 0.560 & 0.451 & 54.36 & 0.417 & -0.901 & - & - & - \\
\bottomrule
\end{tabular}}
\end{table*}

The results support restarting from the original input rather than refining an already altered image. In-place refinement is inferior even to the single-pass setting, with the DF-Score decreasing from $59.6$ to $57.2$. In contrast, redo from the LQ input improves PSNR from $26.01$ to $26.63$\,dB and raises the D-Score/F-Score from $73.0$/$58.5$ to $75.2$/$63.6$, resulting in a DF-Score of $62.0$. This suggests that an unsuccessful intermediate result may contain hallucinated or distorted content that should not be propagated into the next round. Restarting from the LQ input allows the revised instruction to correct the failure without accumulating errors. Importantly, this improvement is not uniform across all samples: only $37\%$ of the 200 test images undergo re-execution, and the gains come from this subset. Images that already produce satisfactory results in the first round remain unchanged. Thus, redo primarily repairs difficult tail cases rather than improving every image uniformly. Detailed statistics on the re-execution frequency and its associated cost are provided in Sec.~\ref{app:cost}.

\subsubsection{VLM Choice}
\label{app:ablation_VLM}

We compare different VLMs for HarnessIR's diagnoser and composer on Real-Paired-200. Verification-driven re-execution is disabled in all variants, allowing us to compare the VLMs without gains from additional execution rounds.

As shown in Tab.~\ref{tab:vlm-ablation}, the results suggest that HarnessIR does not depend on a specific VLM, provided the model is sufficiently capable. The locally deployed Qwen3-VL-32B performs somewhat worse, reaching a DF-Score of $57.5$. GPT-5.6-Luna and Gemini-3.7-Flash perform similarly, with DF-Scores of $59.1$ and $59.6$, respectively. GPT-5.6-Sol achieves the highest PSNR and DF-scores, but its gains over Gemini-3.7-Flash are modest: $0.17$\,dB in PSNR and $0.9$ points in DF-Score. We therefore use Gemini-3.7-Flash in the final system: its restoration performance is comparable, while GPT-5.6-Sol incurs substantially higher API cost (see Sec.~\ref{app:cost}).

\begin{table*}[t]
\centering
\footnotesize
\setlength{\tabcolsep}{3pt}
\caption{Effect of the VLMs used for diagnoser and composer on Real-Paired-200. Verification-driven re-execution is disabled in all variants. Metrics follow Tab.~\ref{tab:mio100_avg}; the best and second-best results in each column are marked in \textcolor{red}{red} and \textcolor{blue}{blue}.}
\label{tab:vlm-ablation}
\vspace{+1mm}
\resizebox{\textwidth}{!}{%
\begin{tabular}{lcccccccccccc}
\toprule
VLM & \multicolumn{4}{c}{Full-reference Metrics} & \multicolumn{5}{c}{No-reference Metrics} & \multicolumn{3}{c}{VLM-based IQA} \\
\cmidrule(lr){2-5}\cmidrule(lr){6-10}\cmidrule(lr){11-13}
 & PSNR$\uparrow$ & SSIM$\uparrow$ & LPIPS$\downarrow$ & DISTS$\downarrow$ & $\Delta$MANIQA$\downarrow$ & $\Delta$CLIP-IQA$\downarrow$ & $\Delta$MUSIQ$\downarrow$ & $\Delta$TOPIQ$\downarrow$ & $\Delta$AFINE-NR$\downarrow$ & D-Score$\uparrow$ & F-Score$\uparrow$ & DF-Score$\uparrow$ \\
\midrule
HarnessIR (Qwen3-VL-32B~\cite{bai2025qwen3}) & 25.68 & 0.760 & 0.267 & 0.161 & 0.034 & \textcolor{blue}{0.024} & 7.82 & 0.067 & 0.050 & 71.2 & 56.1 & 57.5 \\
HarnessIR (GPT-5.6-Sol~\citep{gpt56sol}) & \textcolor{red}{26.18} & \textcolor{red}{0.768} & \textcolor{red}{0.253} & \textcolor{red}{0.151} & \textcolor{red}{0.029} & \textcolor{red}{0.017} & \textcolor{red}{6.48} & \textcolor{red}{0.058} & \textcolor{blue}{0.048} & \textcolor{red}{74.4} & \textcolor{red}{59.7} & \textcolor{red}{60.5} \\
HarnessIR (GPT-5.6-Luna~\citep{gpt56luna}) & \textcolor{blue}{26.05} & 0.762 & \textcolor{blue}{0.259} & \textcolor{blue}{0.155} & 0.034 & 0.027 & 7.78 & 0.066 & 0.059 & 72.7 & \textcolor{blue}{58.7} & 59.1 \\
HarnessIR (Gemini-3.7-Flash~\citep{gemini37}) & 26.01 & \textcolor{blue}{0.763} & 0.261 & 0.158 & \textcolor{blue}{0.032} & \textcolor{blue}{0.024} & \textcolor{blue}{7.72} & \textcolor{blue}{0.062} & \textcolor{red}{0.045} & \textcolor{blue}{73.0} & 58.5 & \textcolor{blue}{59.6} \\
\midrule
GT & - & - & - & - & 0.560 & 0.451 & 54.36 & 0.417 & -0.901 & - & - & - \\
\bottomrule
\end{tabular}}
\end{table*}

\subsection{Cost Analysis}
\label{app:cost}

\begin{table}[t]
\centering
\footnotesize
\setlength{\tabcolsep}{4pt}
\caption{Average per-image runtime, call counts, restoration performance, and cost on Real-Paired-200. VLM and MFM denote API calls. Tool / Perception: Tool reports local model used as restoration or auxiliary tools, perception reports local VLM model used for perception.}
\label{tab:cost}
\vspace{+1mm}
\resizebox{\textwidth}{!}{
\begin{tabular}{lccccccc}
\toprule
Method & VLM & MFM & Tool / Perception & Time (s) & PSNR$\uparrow$ & DF-Score$\uparrow$ & Cost (\$) \\
\midrule
AgenticIR & 0.70 & -- & 5.94 / 9.87 & 88.8 & 22.41 & 38.8 & 0.137 \\
4KAgent & 0.53 & -- & 7.47 / 10.08 & 125.3 & 22.18 & 37.4 & 0.190 \\
\midrule
Baseline (MFM only) & -- & 1.00 & -- & 14.7 & 23.94 & 52.9 & 0.070 \\
HarnessIR (One VLM call) & 1.00 & 1.00 & -- & 27.2 & 25.49 & 56.6 & 0.077 \\ \midrule
HarnessIR (Two VLM calls + Tools) & 2.00 & 1.00 & 2.40 / -- & 41.8 & 26.01 & 59.6 & 0.085 \\ 
HarnessIR (Two VLM calls + Tools, GPT-5.6-Sol) & 2.00 & 1.00 & 2.40 / -- & 45.5 & 26.18 & 60.5 & 0.183 \\ \midrule
HarnessIR (Final, Three VLM calls + Tools + redo) & 3.90 & 1.45 & 2.40 / -- & 64.3 & 26.63 & 62.0 & 0.129 \\
\bottomrule
\end{tabular}}
\end{table}

As shown in Tab.~\ref{tab:cost}, we report the per-image runtime, call counts, restoration performance, and cost of different HarnessIR variants and two previous agentic IR methods~\citep{agenticir,4kagent} on Real-Paired-200. Runtime and VLM call counts come from actual runs. For local computation, including the prior methods' functional modules and HarnessIR's auxiliary tools, we convert measured runtime into cost using on-demand A100 GPU rental rates; API costs are estimated from observed token usage and public list prices. We exclude costs that are difficult to assign to individual images, such as parallel waiting, environment setup, and deployment overhead. Rental rates provide the fairest method-independent basis for comparison, whereas GPU purchase costs depend on assumptions about depreciation, utilization, service lifetime, and allocation, making per-image costs ill-defined and incomparable. The resulting values should therefore be viewed as \textbf{idealized references} for computational and monetary efficiency. We discuss the results first and detail the calculations afterward.

\vspace{2mm} \noindent\textbf{Results Analysis.}
Compared with the two previous agentic IR methods, the final HarnessIR system achieves a higher DF-Score on Real-Paired-200 at a lower estimated per-image cost: $\$0.129$, compared with $\$0.137$ for AgenticIR and $\$0.190$ for 4KAgent. For the HarnessIR variants, starting from direct MFM usage, adding a single VLM call raises the DF-Score from $52.9$ to $56.6$ and the PSNR from $23.94$ to $25.49$\,dB, while increasing the cost only from $\$0.070$ to $\$0.077$ per image. This one-call variant requires no local GPU computation, making it a particularly attractive option when simplicity and deployment cost are priorities. Adding a second VLM call and local auxiliary tools raises the cost modestly to $\$0.085$, while further improving the DF-Score to $59.6$ and the PSNR to $26.01$\,dB. Thus, the diagnosis-and-composition pipeline provides a favorable additional quality gain for a small increase in cost.

The complete HarnessIR system adds verification and selective re-execution, reaching a DF-Score of $62.0$ and a PSNR of $26.63$\,dB at an average cost of $\$0.129$ per image. This average gain is driven by the subset of difficult cases that trigger re-execution rather than by uniformly improving every image: 37\% of images undergo at least one redo, and the verifier retains the better candidate. The increased cost, therefore, supports targeted correction of cases that need it. Finally, replacing Gemini-3.7-Flash with GPT-5.6-Sol in the two-VLM configuration increases the cost from $\$0.085$ to $\$0.183$ per image. Although GPT-5.6-Sol yields some performance improvement, the gain is modest relative to the more than twofold cost increase. We therefore use Gemini-3.7-Flash in the final system to balance performance and cost.

\vspace{2mm} \noindent\textbf{Calculation Details.}
The GPU price is the mean on-demand A100 rate of the three major cloud providers, Google Cloud \footnote{GCP a2-ultragpu-1g: \url{https://cloud.google.com/products/compute/pricing/accelerator-optimized}}, Microsoft Azure \footnote{Azure ND96asr\_A100\_v4: \url{https://azure.microsoft.com/pricing/details/virtual-machines/linux/}}, and Amazon Web Services\footnote{AWS p4de.24xlarge: \url{https://aws.amazon.com/cn/savingsplans/compute-pricing/}}, normalized to a per-GPU hourly rate: \$2.671 per GPU-hour. All locally executed calls, including local restoration, perception, and auxiliary-tool calls, are converted to GPU-rental cost using this rate. Following the deployment setup of 4KAgent, both AgenticIR and 4KAgent use two GPUs: one GPU is dedicated to DepictQA, while the other runs the largest model so that it remains within the available GPU memory. For AgenticIR and 4KAgent, call counts and runtimes are averaged over Real-Paired-200.

We calculate API costs using public list prices, retrieved in September 2026, without cached-input or batch discounts.\footnote{Gemini and Nano~Banana~2 pricing: \url{https://ai.google.dev/gemini-api/docs/pricing}. GPT pricing: \url{https://openai.com/api/pricing/}.} The relevant prices are reported as input/output rates per million tokens: Nano~Banana~2 uses in/out rates of \$0.50/\$60 for image output and \$0.50/\$3 for text output; Gemini-3.7-Flash uses an in/out rate of \$0.75/\$3.75; and GPT-5.6-Sol uses an in/out rate of \$4/\$20. Thus, GPT-5.6-Sol is more than five times as expensive as Gemini-3.7-Flash for both input and output tokens. For HarnessIR, API costs are calculated from the actual input and output usage of the VLM and MFM calls, while the cost of local auxiliary tools is calculated through GPU-rental time. Under HarnessIR-NB2, one Gemini-3.7-Flash VLM call costs approximately \$0.007 (different under different token counts), one MFM call costs approximately \$0.070, and one complete auxiliary-tool suite costs approximately \$0.0048. In the final system, every image undergoes diagnosis, composition, execution, and verification. About 29\% of the 200 images trigger one redo and 8\% trigger two redos; thus, 37\% of the 200 images trigger at least one redo, corresponding to an average of $0.45$ redo rounds per image. Each redo adds two VLM calls and one MFM call, yielding $3.90$ VLM calls and $1.45$ MFM calls per image on average. The costs reported in Tab.~\ref{tab:cost} are computed from unrounded per-image token usage and runtime measurements.

\begin{figure}[t]
  \centering
  \includegraphics[width=0.52\linewidth]{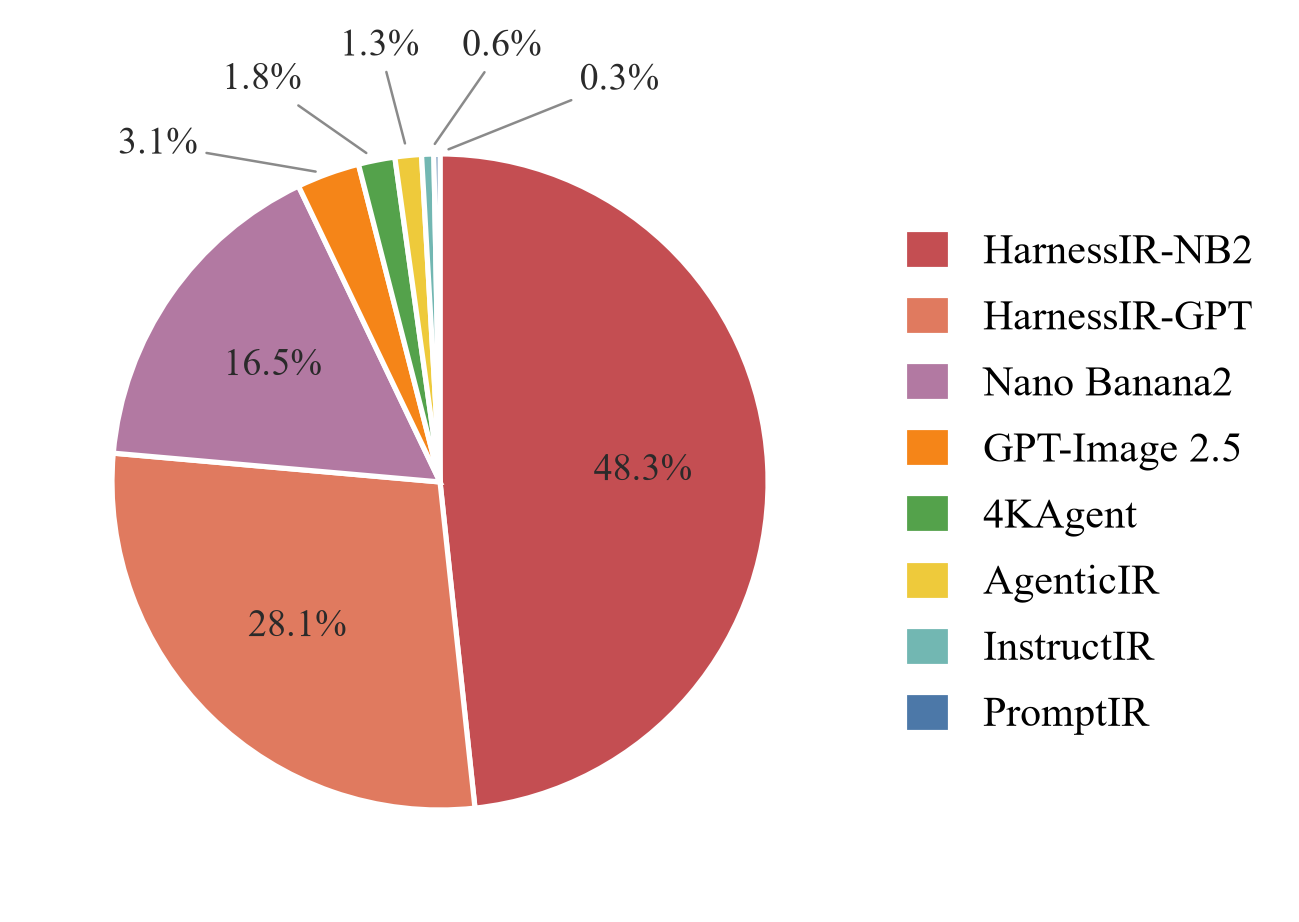}
  \vspace{-8mm}
  \caption{Human selection rates among our methods and baselines.}
  \label{fig:user-study}
\end{figure}

\subsection{User Study}
\label{app:userstudy}

We conduct a user study to compare restoration quality on real images.
For each test case, annotators see one low-quality (LQ) image together with the eight restorations of that image. The eight methods are the same set used in the DF-Score comparison: PromptIR \citep{PromptIR}, InstructIR \citep{InstructIR}, 4KAgent \citep{4kagent}, AgenticIR \citep{agenticir}, GPT-Image~2.5 \citep{gptimage25}, Nano Banana 2 \citep{nanobanana2}, HarnessIR-GPT and HarnessIR-NB2. The eight results are shown in random order. Method names and metric values are withheld as well. Annotators are required to select the single restoration they judge to be the best. The instruction asks them to prefer a result that removes the target degradation, preserves the main scene content, and avoids obvious artifacts.

We draw 100 images at random from the Real-Paired-200 and 100 images at random from Real-NoGT-200, giving 200 groups in total. Each of the 25 participants annotates all 200 groups, so each group receives 25 independent votes and the study contains $25\times200=5000$ votes.

The selection rate of a method is the percentage of the 5000 votes in which annotators choose that method as the best restoration. Fig.~\ref{fig:user-study} shows a clear preference for the harnessed restorations. HarnessIR-NB2 receives $48.3\%$ of the votes and HarnessIR-GPT receives $28.1\%$, so the two HarnessIR variants together account for $76.4\%$. These human preferences are consistent with the quantitative results and qualitative comparisons presented above. They further confirm that HarnessIR substantially improves restoration quality while maintaining high perceptual quality and faithfully preserving the original content of the LQ images.

\section{Conclusion}

We presented \textbf{HarnessIR}, an agentic framework that used MFMs as the executor for universal real-world IR. By combining image-specific diagnosis, auxiliary tools, prompt composition, and verification-driven re-execution, HarnessIR harnessed the MFM to remove degradations without altering image content. Across synthetic and challenging real-world benchmarks, HarnessIR consistently improved restoration performance over the direct use of MFM, achieving particularly large gains on real-world images. Our ablations further showed that these improvements came from the coordinated design of the harness at an acceptable cost. Overall, HarnessIR demonstrated that a carefully designed harness can substantially unlock the restoration potential of proper MFMs and turn them into more effective and controllable universal real-world restorers.

\vspace{2mm} \noindent\textbf{Limitations.}
HarnessIR requires access to online VLM and MFM APIs, which is a common requirement by existing agentic IR methods such as AgenticIR \citep{agenticir} and 4KAgent \citep{4kagent} (using GPT-4o). Moreover, an agentic harness can only elicit and regulate the capabilities of its executor but cannot raise the executor's performance ceiling. In the future, with this general IR agent, we can curate diverse and challenging restoration data and generate verification-guided training supervision to improve the capabilities of local and smaller models.



{\small
\bibliography{ref}
}

\appendix

\clearpage
\setcounter{table}{0}
\setcounter{equation}{0}
\setcounter{figure}{0}
\renewcommand{\thetable}{\thesection.\arabic{table}}
\renewcommand{\theequation}{\thesection.\arabic{equation}}
\renewcommand{\thefigure}{\thesection.\arabic{figure}}
\begin{center}
    {\LARGE\bfseries Appendix}
\end{center}

In this appendix, we provide the following materials:
\begin{itemize}

    \item \textbf{A. Input Request and A Complete Case Study}: The fixed request template for each degradation category to represent the usage of general users and a real example processed by HarnessIR.

    \item \textbf{B. Real Test Set Composition}: Detailed construction and data sources of Real-Paired-200 and Real-NoGT-200.

    \item \textbf{C. DF-Score Evaluation Protocol}: The alignment with human judgments, and evaluator-VLM replacement experiments.
    
    \item \textbf{D. Detailed Results on MiO100}: Per-group results for MiO100 Groups A, B, and C.

    \item \textbf{E. More Visual Results}: Additional qualitative comparisons between HarnessIR and other representative methods.

\end{itemize}

\section{Input Request and A Complete Case Study}

\subsection{Input Request}
\label{app:Request}

We use short, fixed requests to represent the intent of a general user. The request for \emph{mix} serves as the default. When one or more degradation types are specified, we prepend a removal instruction to the same default request. For mixed degradations, the names of all applicable degradations are joined with ``and'' or followed in the end. For example, an image containing both haze and rain needs to be brightened is assigned the request ``Please remove the haze and rain from the image and restore this low-quality image to a clear, clean, and natural state.  Meanwhile, recover its normal brightness''. The basic requests are as follows:

\begin{lstlisting}[style=hirprompt]
### [[mix]] ###
Restore this low-quality image to a clear, clean and natural state.

### [[haze]] ###
Please remove the haze from the image and restore this low-quality image to a clear, clean and natural state.

### [[rain]] ###
Please remove the rain from the image and restore this low-quality image to a clear, clean and natural state.

### [[snow]] ###
Please remove the snow from the image and restore this low-quality image to a clear, clean and natural state.

### [[lowlight]] ###
Restore this low-quality image to a clear, clean and natural state and recover its normal brightness.

\end{lstlisting}

\subsection{A Complete Case Study}
\label{app:completecase}

\begin{figure*}[t]
    \centering
        \includegraphics[width=\linewidth]{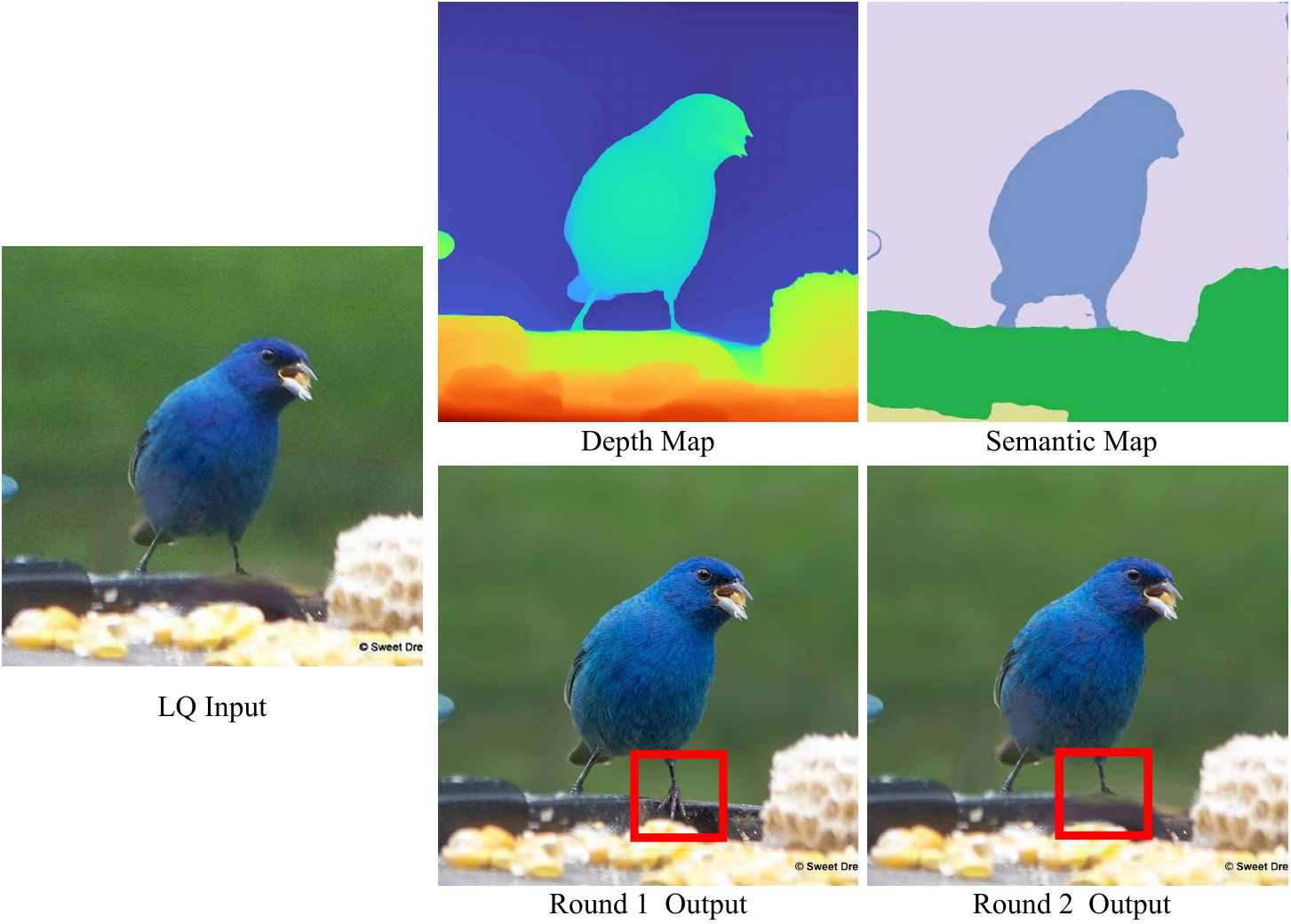}
        \vspace{-5mm}
        \caption{A complete case study illustrating the workflow of HarnessIR, including the input image, depth map and semantic map from auxiliary tools, and the round1 and round2 output images.}
        \label{fig:case-study}
    \vspace{-2mm}
\end{figure*}

To make the workflow concrete, Fig.~\ref{fig:case-study} presents a real example processed by HarnessIR. Below, we provide the corresponding outputs at each stage: the initial diagnosis, the evidence returned by the selected auxiliary tools, the instruction composed for the MFM executor, and the verifier's decision on whether to re-execute. This example illustrates how image-specific observations are translated into restoration requirements and how the resulting image is checked against the original input.

\vspace{2mm}
\noindent\textbf{Output of Perception and Diagnosis}. First, the input to HarnessIR contains the LQ image (left part of Fig.~\ref{fig:case-study}) and the user request ``Restore this low-quality image to a clear, clean and natural state." The output of the diagnoser is as follows:

\begin{lstlisting}[style=hirprompt]
  "triage": {
    "content": "A bright blue bird holding a seed in its beak, standing on a dark feeder. The foreground contains scattered yellow corn kernels and a white textured block, both heavily out of focus. The background is a smooth, uniform green, indicating shallow depth of field in daylight.",
    "degradations": [
      {
        "type": "noise",
        "severity": "moderate",
        "spatial": "global",
        "location": "whole image",
        "worst_at": "green background and dark blue shadows on the bird",
        "mildest_at": "bright white areas in foreground",
        "evidence": "grainy texture visible across smooth surfaces"
      },
      {
        "type": "softness",
        "severity": "moderate",
        "spatial": "local",
        "location": "midground (bird)",
        "worst_at": "bird's body feathers",
        "mildest_at": "bird's eye and beak",
        "evidence": "lack of fine feather detail despite being the focal point"
      },
      {
        "type": "defocus-blur",
        "severity": "severe",
        "spatial": "local",
        "location": "foreground and background",
        "worst_at": "background green and foreground white block",
        "mildest_at": "midground bird",
        "evidence": "optical shallow depth of field blurring non-focal planes"
      },
      {
        "type": "jpeg",
        "severity": "mild",
        "spatial": "global",
        "location": "whole image",
        "worst_at": "edges of the bird against the green background",
        "mildest_at": "uniform background areas",
        "evidence": "slight blockiness and ringing around high-contrast boundaries"
      }
    ],
    "sensitive": {
      "has_text": true,
      "has_faces": false,
      "has_layered_depth": true,
      "has_complex_semantics": true,
      "notes": "Fine feather textures on the bird."
    },
    "tool_plan": [
      {
        "tool": "T1_OCR",
        "reason": "The watermark text in the bottom right needs to be preserved and not distorted by restoration."
      },
      {
        "tool": "T3_DEPTH",
        "reason": "The image relies heavily on shallow depth of field; a depth map will help ensure the foreground and background remain soft while the midground bird is sharpened."
      },
      {
        "tool": "T4_SEGMENTATION",
        "reason": "Separating the bird from the background will allow for targeted detail enhancement on the feathers without amplifying noise in the smooth green bokeh."
      }
    ],
    "difficulty_hint": "medium",
    "difficulty_reason": "Recovering fine feather detail on the slightly soft bird without over-sharpening the intentionally blurred foreground and background requires careful spatial control."
  }

\end{lstlisting}

The diagnosis identifies text, layered depth, and semantically distinct regions in the image. Therefore, it requests OCR, depth estimation, and semantic segmentation tools to gather image-specific evidence for restoration.

\vspace{2mm}
\noindent\textbf{Output of On-Demand Tool Call}. The selected auxiliary tools return the OCR transcription, depth map, and semantic segmentation map shown in Fig.~\ref{fig:case-study}. The OCR result is as follows:

\begin{lstlisting}[style=hirprompt]
  "tool_evidence": {
    "ocr": {
      "engine": "ppocr-v6",
      "n_texts": 1,
      "texts": [
        {
          "string": "© Sweet Dre",
          "confidence": 0.9435150027275085,
          "bbox_norm": [
            0.839,
            0.932,
            1.0,
            0.97]}]}
\end{lstlisting}

In this example, there is no face, so there is no output related to face detection tool. 

\vspace{2mm}
\noindent\textbf{Output of Prompt Composition}. The composer then converts the diagnosis and tool evidence into a restoration instruction for the MFM, as shown below. In particular, it uses the depth information to identify the bird and feeder as the intended focal plane, while recognizing the blur in the foreground and background as a natural shallow-depth-of-field effect that should be preserved rather than restored. The prompt also describes the scene and the bird's recoverable details, and records the text in the lower-right corner so that it can be preserved unchanged.

\begin{lstlisting}[style=hirprompt]

You are an image restoration expert. Restore this photograph.\n\nThe image shows a bright blue bird holding a seed in its beak, standing on a dark feeder. The foreground contains out-of-focus corn kernels and a suet block, set against a smooth, out-of-focus green background.\n\nFidelity is your first priority. Preserve the exact scene content, geometry, object positions, and structure. Do not add, remove, or alter any objects. Maintain the current global color, tone, and white balance exactly as they are; do not shift hues or recolor the image.\n\nCleanly remove sensor noise, grain, and JPEG compression artifacts (such as blockiness and ringing around high-contrast edges) across the entire image. Ensure no over-smoothing erases genuine textures, and avoid introducing over-sharpening halos, ghosting, or visible seams.\n\nThe bird and the dark feeder it stands on represent the intended focal plane but suffer from capture softness. Genuinely deblur this midground region. Recover fine, crisp detail in the bird's feathers, eye, beak, and the seed it holds. The bird must be clearly resolved and textured without looking artificially sharpened.\n\nThe shallow depth of field is a physical property of this photograph. The out-of-focus yellow corn and white block in the foreground, as well as the smooth green background, must remain exactly as out of focus as they currently are. Do not attempt to sharpen or deblur these areas; simply denoise them so the optical bokeh is clean and smooth.\n\nFinally, in the bottom right corner, preserve the watermark text exactly as it appears. The characters \"© Sweet Dre\" must remain legible and unchanged; do not invent, guess, or alter any letters.

\end{lstlisting}

\vspace{2mm}
\noindent\textbf{Output of Verification and Re-Execution}.
After execution, the generated image is passed to the verifier, whose response indicates that re-execution is necessary. As shown in the first-round result (see Fig.~\ref{fig:case-study}), one readily identifiable issue is that the bird's left claw should be occluded in the input image, but it is incorrectly generated in the restoration result. The Verifier detects this error, and the revised instruction explicitly emphasizes that the occluded claw should not be reconstructed. Correspondingly, the second-round output of Fig.~\ref{fig:case-study} no longer exhibits this problem. In the next round of verification, this result shows that there is no need to redo it.

\begin{lstlisting}[style=hirprompt]
    {
        "reasoning": "The input suffers from moderate blur, noise, and compression artifacts, obscuring fine details like the bird's feathers. The attempt aggressively denoises and sharpens the image, but goes too far in adding detail. Tracing back to Picture 1, the left claw of the bird in the picture should have been obscured, but now the error has occurred and it has been placed in front of the dark feeder. The crisp, distinct feathers on the bird's back and breast in the attempt are not supported by the blurry input; they are fabricated. Additionally, there is an inconsistent, smooth, blurry patch on the bird's lower left flank where the model failed to generate texture, contrasting sharply with the hallucinated details elsewhere. The decisive issue is the hallucination of high-frequency feather details.",
        "checklist": {
          "U1": "fail",
          "U2": "fail",
          "U3": "pass",
          "U4": "na",
          "U5": "pass",
          "U6": "pass",
          "U7": "fail",
          "U8": "pass",
          "U9": "fail",
          "U10": "fail",
          "T1": "pass",
          "T2": "pass",
          "T3": "fail",
          "T4": "fail",
          "T5": "na",
          "T6": "na",
          "T7": "pass",
          "T8": "fail"
        },
        "defects": [
           {
            "location": "bird's left claw",
            "bbox_norm": [
              0.45,
              0.67,
              0.69,
              0.90
            ],
            "issue": "Compared with the input, it should be obscured.",
            "item": "U1",
            "severity": "moderate"
          },
          {
            "location": "bird's back and breast",
            "bbox_norm": [
              0.35,
              0.3,
              0.65,
              0.6
            ],
            "issue": "Fabricated feather textures not supported by the blurry input.",
            "item": "U2",
            "severity": "severe"
          },
          {
            "location": "bird's lower left flank",
            "bbox_norm": [
              0.31,
              0.53,
              0.41,
              0.68
            ],
            "issue": "Inconsistent smooth, blurry patch lacking texture.",
            "item": "U9",
            "severity": "moderate"
          }
        ],
        "scores": {
          "restoration_sufficiency": 6,
          "fidelity": 3,
          "scope_discipline": 3,
          "naturalness": 4
        },
        "overall": 3.0,
        "needs_redo": true,
        "redo_priority": 8.0,
        "best_round": 0,
        "best_round_why": "only one attempt",
    },
\end{lstlisting}

The revised instruction from the composer is as follows:

\begin{lstlisting}[style=hirprompt]
   "refine": {
        "mode": "global",
        "framing": "defect",
        "new_instruction": "You are an image restoration expert. Restore this photograph.\n\nThe image shows a bright blue bird holding a seed in its beak, standing on a dark feeder. The foreground contains out-of-focus corn kernels and a suet block, set against a smooth, out-of-focus green background.\n\nFidelity is your first priority. Preserve the exact scene content, geometry, object positions, and structure. Do not add, remove, or alter any objects. Maintain the current global color, tone, and white balance exactly as they are; do not shift hues or recolor the image.\n\nCleanly remove sensor noise, grain, and JPEG compression artifacts (such as blockiness and ringing around high-contrast edges) across the entire image. Ensure no over-smoothing erases genuine textures, and avoid introducing over-sharpening halos, ghosting, or visible seams.\n\nThe bird and the dark feeder it stands on represent the intended focal plane but suffer from capture softness. Genuinely deblur this midground region to recover realistic detail in the bird's eye, beak, and the seed it holds. However, the left claw of the bird in the picture should have been obscured, don't try to generate it. You must be extremely careful with the bird's plumage. Do not invent or hallucinate \"fabricated feather textures not supported by the blurry input\" on the bird's back and breast. Any texture enhancement must be strictly derived from existing pixel data. Furthermore, the texture recovery must be consistent across the entire body of the bird. Do not leave an \"inconsistent smooth, blurry patch lacking texture\" on the bird's lower left flank while sharpening the rest of the body. Ensure the level of detail is uniform, balanced, and physically plausible based only on the underlying image data.\n\nThe shallow depth of field is a physical property of this photograph. The out-of-focus yellow corn and white block in the foreground, as well as the smooth green background, must remain exactly as out of focus as they currently are. Do not attempt to sharpen or deblur these areas; simply denoise them so the optical bokeh is clean and smooth.\n\nFinally, in the bottom right corner, preserve the watermark text exactly as it appears. The characters \"© Sweet Dre\" must remain legible and unchanged; do not invent, guess, or alter any letters.",
\end{lstlisting}

\section{Real Test Set Composition}
\label{app:real_datasets}

\begin{table}[t]
\centering
\footnotesize
\setlength{\tabcolsep}{4pt}
\caption{Source datasets of the two real-world test sets, with the number of
images contributed by each in parentheses.}
\vspace{+1mm}
\label{tab:real_sources}
\resizebox{\textwidth}{!}{
\begin{tabular}{llr}
\toprule
Degradation & Source datasets (images) & Total \\
\midrule
\multicolumn{3}{l}{\emph{Real-Paired-200} --- 200 LQ/GT pairs} \\
\midrule
Low resolution & DRealSR~(20)~\citep{drealsr}, RealSR~(20)~\citep{realsr} & 40 \\
Haze & I-HAZE~(20)~\citep{ihaze}, NH-HAZE~(20)~\citep{nhhaze} & 40 \\
Rain & RealRain-1k-H~(20)~\citep{realrain}, RainDS-real~(20)~\citep{rainds} & 40 \\
Blur & RealBlur~(20)~\citep{rim2020realblur} & 20 \\
Sensor noise & SIDD~(20)~\citep{sidd}, PolyU~(20)~\citep{polyunoise} & 40 \\
Mixed, in the wild & RAIM~(20)~\citep{raim} & 20 \\
\midrule
\multicolumn{3}{l}{\emph{Real-NoGT-200} --- 200 LQ images, no reference} \\
\midrule
Mixed, no named weather & RealDeg~(classic film (25),old photo(25) and social media(25))~\citep{chen2024faithdiff}, ImageNet-worst~(25)~\citep{deng2009imagenet} & 100 \\
Haze & ACDC-fog~(7)~\citep{acdc}, JarvisIR-fog~(6)~\citep{jarvisir}, RESIDE-RTTS~(6)~\citep{RESIDE}, NightHaze~(6)~\citep{nighthaze} & 25 \\
Rain & ACDC-rain~(7)~\citep{acdc}, HQ-NightRain-Real~(6)~\citep{hqnightrain}, JarvisIR-rain~(6)~\citep{jarvisir}, NTURain-real~(6)~\citep{chen2018robust} & 25 \\
Snow & ACDC-snow~(9)~\citep{acdc}, JarvisIR-snow~(8)~\citep{jarvisir}, Snow100K-realistic~(8)~\citep{desnownet} & 25 \\
Low light & ExDark~(9)~\citep{Exdark}, ACDC-night~(8)~\citep{acdc}, JarvisIR-night~(8)~\citep{jarvisir} & 25 \\
\bottomrule
\end{tabular}}
\end{table}

We construct two real-world test sets from publicly available datasets: Real-Paired-200, which contains 200 paired low-quality and ground-truth images, and Real-NoGT-200, which contains 200 real-world low-quality images without ground-truth. The source datasets and the number of selected images are summarized in Tab.~\ref{tab:real_sources}. We randomly sample images from each source dataset; the number selected from each is given in parentheses in Tab.~\ref{tab:real_sources}. For consistency across datasets, images whose longer side is shorter than 1024 pixels are isotropically resized so that their longer side becomes 1024 pixels. Images whose longer side exceeds 1024 pixels are center-cropped along the longer dimension until it is 1024 pixels, while the shorter dimension is left unchanged; images whose longer side is already 1024 pixels are left unchanged. For paired samples, the corresponding ground-truth image undergoes the matching resize or center-crop operation so that it has the same final spatial resolution and remains pixel-aligned with the low-quality image.

\section{DF-Score Evaluation Protocol}
\label{app:dfscore}

This section describes the evaluation protocol and reliability analysis of DF-Score. First, since the system prompt used by the independent VLM evaluator to assign D-Score and F-Score is several pages long, we provide it in released codes. We then assess the alignment of DF-Score and representative no-reference IQA metrics with human judgments through a pairwise study. Finally, we test whether D-Score, F-Score, and DF-Score remain consistent across VLM evaluators from different model families.

\vspace{2mm} \noindent\textbf{Alignment with Human Judgments.}
We evaluate how well DF-Score and commonly used NR-IQA metrics agree with human judgments of restoration quality. We construct 300 comparison pairs from the Real-NoGT-200 test set. Each pair contains one LQ image and two restorations of that image produced by distinct methods. The method pool includes PromptIR \citep{PromptIR}, InstructIR \citep{InstructIR}, 4KAgent \citep{4kagent}, AgenticIR \citep{agenticir}, GPT-Image~2.5 \citep{gptimage25}, Nano Banana~2 \citep{nanobanana2}, HarnessIR-GPT, and HarnessIR-NB2. For each pair, two methods are sampled uniformly from this pool. The left--right order of the restorations is randomized, and method identities and metric scores are hidden from the annotators.

We recruit $30$ annotators. For each pair, annotators are shown the LQ image together with the two restorations and asked to select the better result. The instructions emphasize that a preferable restoration should remove the target degradation, preserve the content and structure of the input image, and avoid visible artifacts. Each annotator provides one forced-choice judgment per pair.

\begin{figure}[t]
  \centering
  \includegraphics[width=0.70\linewidth]{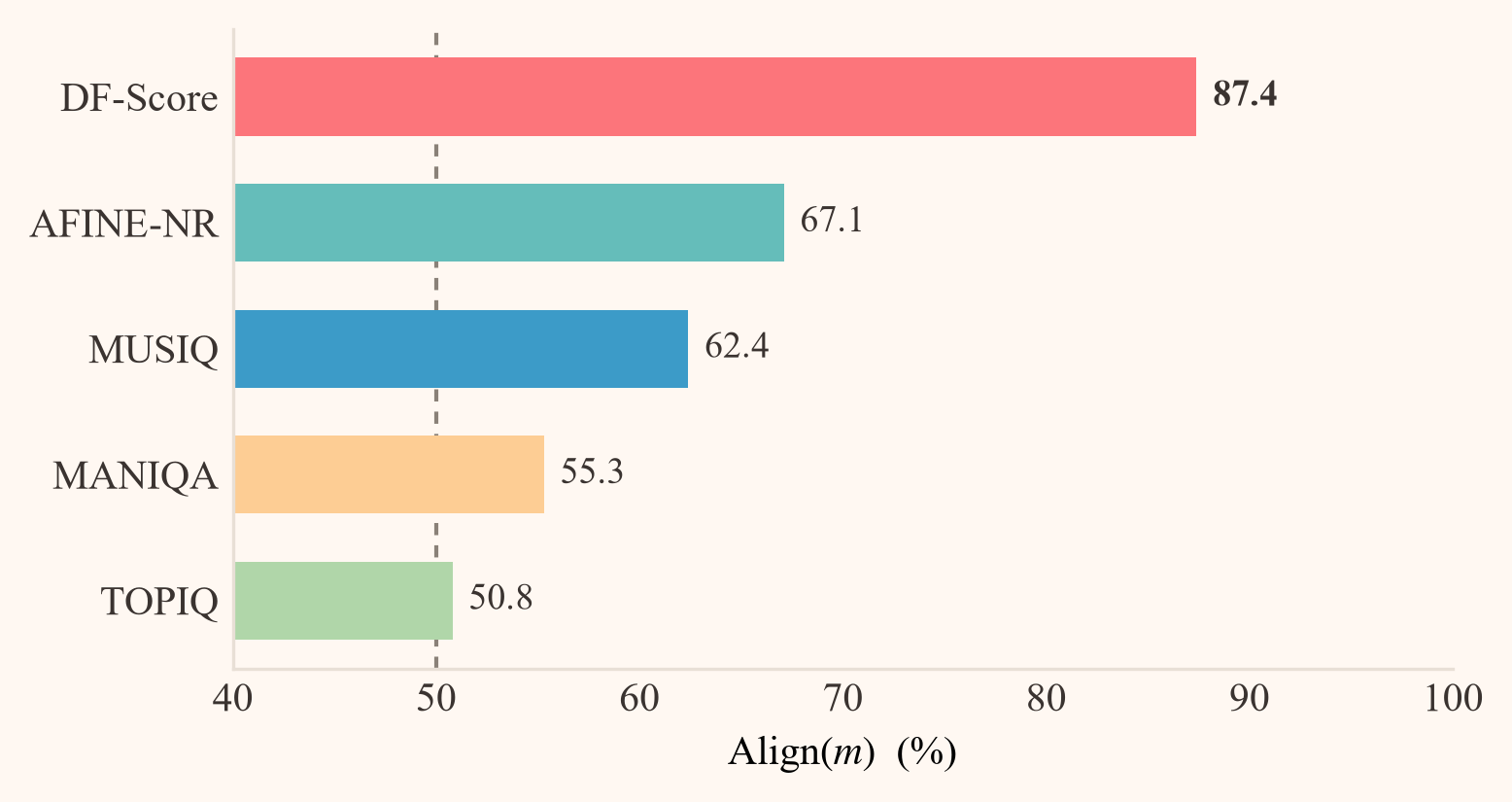}
  \caption{Pairwise $\mathrm{Align}(m)$ with majority human preference on the real no-GT test set. Dashed line: chance ($50\%$). }
  \label{fig:df-human}
\end{figure}

We aggregate the annotations by majority vote to obtain the human preference for each pair. Pairs without a strict majority are discarded. For a metric $m$, we measure its agreement with human judgments using the alignment ratio:
\[
\mathrm{Align}(m)
=
\frac{
\#\{\text{valid pairs whose metric ordering agrees with the majority vote}\}
}{
\#\{\text{valid pairs}\}
}.
\]
A random ranking gives an expected alignment of $50\%$. We report the alignment of DF-Score and four representative NR-IQA metrics---MUSIQ, AFINE-NR, TOPIQ, and MANIQA---in Fig.~\ref{fig:df-human}. DF-Score agrees with human preference in $87\%$ of valid comparisons, whereas the NR-IQA metrics achieve lower alignment rates, ranging from $50.8\%$ to $67.1\%$. This suggests that, when ground-truth images are unavailable, DF-Score may better reflect human preferences than the NR-IQA metrics.

\vspace{2mm} \noindent\textbf{Consistency Across Evaluator Families.}
We examine whether D-Score, F-Score, and their geometric mean, DF-Score, depend strongly on the choice of VLM evaluator. Using the restored outputs for the 400 images in Real-Paired-200 and Real-NoGT-200, we obtain scores from GPT-5.6-Sol and Claude-Opus-5 under the same evaluation protocol and compute the Pearson linear correlation coefficient (PLCC) between each evaluator's scores and those of Gemini-3.7-Flash. Tab.~\ref{tab:evaluator_consistency} reports the correlations for all three scores. Despite the presence of some outliers, the correlations are consistently high across all three measures, ranging from $0.85$ to $0.94$. This result suggests that DF-Score is robust to the choice of VLM evaluator. Assessing whether degradations have been removed and whether the original image content has been preserved is comparatively straightforward, and different VLMs appear capable of making these judgments consistently, even though their absolute scores may differ.

\begin{table}[t]
\centering
\caption{PLCC of D-Score, F-Score, and DF-Score between each alternative VLM evaluator and Gemini-3.7-Flash on restored outputs for the 400 images in Real-Paired-200 and Real-NoGT-200.}
\label{tab:evaluator_consistency}
\vspace{-1mm}
\small
\begin{tabular}{lccc}
\toprule
Evaluator & \multicolumn{3}{c}{PLCC vs. Gemini-3.7-Flash~\citep{gemini37}} \\
\cmidrule(lr){2-4}
 & D-Score & F-Score & DF-Score \\
\midrule
GPT-5.6-Sol~\citep{gpt56sol} & 0.85 & 0.87 & 0.94 \\
Claude-Opus-5~\citep{claude-opus-5} & 0.87 & 0.89 & 0.92 \\
\bottomrule
\end{tabular}
\vspace{-2mm}
\end{table}

\section{Detailed Results on MiO100}
\label{app:mio100}

\begin{table}[t]
\centering
\caption{Per-group results on MiO100, for which Tab.~\ref{tab:mio100_avg} reports the average. Marking and metrics follow Tab.~\ref{tab:mio100_avg}.}
\label{tab:mio100_abc}
\vspace{+1mm}
\footnotesize
\setlength{\tabcolsep}{3pt}
\resizebox{\textwidth}{!}{%
\begin{tabular}{llcccccccccccc}
\toprule
Datasets & Method & \multicolumn{4}{c}{Full-reference Metrics} & \multicolumn{5}{c}{No-reference Metrics} & \multicolumn{3}{c}{VLM-based IQA} \\
\cmidrule(lr){3-6}\cmidrule(lr){7-11}\cmidrule(lr){12-14}
 & & PSNR$\uparrow$ & SSIM$\uparrow$ & LPIPS$\downarrow$ & DISTS$\downarrow$ & $\Delta$MANIQA$\downarrow$ & $\Delta$CLIP-IQA$\downarrow$ & $\Delta$MUSIQ$\downarrow$ & $\Delta$TOPIQ$\downarrow$ & $\Delta$AFINE-NR$\downarrow$ & D-Score$\uparrow$ & F-Score$\uparrow$ & DF-Score$\uparrow$ \\
\midrule
\multirow{18}{*}{Group A}
 & AirNet & 20.70 & 0.66 & 0.43 & 0.24 & 0.19 & 0.27 & 26.37 & 0.31 & 0.25 & 20.6 & 88.3 & 22.42 \\
 & PromptIR & 21.29 & 0.67 & 0.43 & 0.25 & 0.19 & 0.26 & 26.14 & 0.31 & 0.24 & 21.6 & \textcolor{red}{92.9} & 24.57 \\
 & MiOIR & 21.74 & 0.69 & 0.39 & 0.23 & 0.19 & 0.27 & 23.93 & 0.30 & 0.23 & 27.5 & \textcolor{blue}{89.1} & 35.04 \\
 & DA-CLIP & 20.87 & 0.65 & 0.42 & 0.24 & 0.19 & 0.26 & 24.94 & 0.30 & 0.24 & 20.4 & \textcolor{red}{92.9} & 25.87 \\
 & InstructIR & 21.52 & 0.64 & 0.47 & 0.25 & 0.21 & 0.31 & 27.86 & 0.34 & 0.27 & 24.4 & 87.9 & 31.96 \\
 & AutoDIR & 21.04 & 0.67 & 0.37 & 0.22 & 0.12 & 0.24 & 15.18 & 0.24 & 0.14 & 40.1 & 66.2 & 38.11 \\
 \cmidrule(l){2-14}
 & AgenticIR & 21.53 & \textcolor{blue}{0.70} & 0.31 & 0.19 & 0.11 & 0.17 & 8.80 & 0.17 & 0.07 & 54.8 & 47.1 & 48.09 \\
 & MAIR & 21.02 & 0.67 & 0.30 & - & 0.31 & 0.17 & 9.47 & - & - & - & - & - \\
 & 4KAgent & 21.48 & 0.67 & 0.30 & 0.18 & 0.08 & 0.10 & 5.47 & 0.11 & 0.08 & 49.4 & 32.7 & 36.78 \\
 & TIR-Agent & 22.07 & 0.69 & 0.29 & - & 0.25 & 0.10 & 4.97 & - & - & - & - & - \\
 & SEAR & 21.80 & \textcolor{blue}{0.70} & 0.30 & - & 0.30 & 0.15 & 7.37 & - & - & - & - & - \\
 & OPERA & \textcolor{blue}{22.32} & \textcolor{red}{0.71} & 0.33 & - & 0.29 & 0.20 & 9.32 & - & - & - & - & - \\
 \cmidrule(l){2-14}
 & GPT-Image-2.5 & 17.48 & 0.53 & 0.37 & 0.20 & \textcolor{blue}{0.02} & 0.11 & 4.78 & 0.07 & \textcolor{red}{0.00} & 42.8 & 20.1 & 27.52 \\
 & HarnessIR-GPT & \textbf{20.19} & \textbf{0.61} & \textbf{0.28} & \textbf{0.16} & \textcolor{blue}{\textbf{0.02}} & \textbf{0.06} & \textbf{2.87} & \textcolor{red}{\textbf{0.02}} & \textcolor{red}{\textbf{0.00}} & \textbf{71.2} & \textbf{45.5} & \textbf{54.81} \\
 \cmidrule(l){2-14}
 & Nano Banana 2 & 21.38 & 0.66 & \textcolor{blue}{0.22} & \textcolor{blue}{0.13} & \textcolor{red}{0.00} & \textcolor{blue}{0.04} & \textcolor{blue}{0.35} & \textcolor{blue}{0.04} & \textcolor{blue}{0.01} & \textcolor{red}{80.2} & 51.4 & \textcolor{blue}{61.75} \\
 & HarnessIR-NB2 & \textcolor{red}{\textbf{22.45}} & \textbf{0.67} & \textcolor{red}{\textbf{0.20}} & \textcolor{red}{\textbf{0.12}} & \textcolor{red}{\textbf{0.00}} & \textcolor{red}{\textbf{0.03}} & \textcolor{red}{\textbf{0.31}} & \textcolor{blue}{\textbf{0.04}} & \textcolor{red}{\textbf{0.00}} & \textcolor{blue}{78.8} & \textbf{58.5} & \textcolor{red}{\textbf{65.47}} \\
 \cmidrule(l){2-14}
 & GT & - & - & - & - & 0.64 & 0.65 & 68.66 & 0.64 & -0.91 & - & - & - \\
 & LQ & 20.20 & 0.61 & 0.50 & 0.28 & 0.21 & 0.28 & 29.06 & 0.34 & 0.27 \\
\midrule
\multirow{18}{*}{Group B}
 & AirNet & 19.97 & 0.70 & 0.38 & 0.22 & 0.16 & 0.23 & 21.46 & 0.25 & 0.20 & 28.8 & 83.1 & 33.91 \\
 & PromptIR & 20.38 & 0.71 & 0.36 & 0.22 & 0.15 & 0.23 & 21.42 & 0.25 & 0.19 & 31.6 & 88.1 & 35.52 \\
 & MiOIR & 21.19 & 0.72 & 0.34 & 0.21 & 0.17 & 0.22 & 20.66 & 0.27 & 0.21 & 39.0 & \textcolor{red}{90.5} & 48.97 \\
 & DA-CLIP & 20.76 & 0.69 & 0.38 & 0.23 & 0.16 & 0.26 & 20.91 & 0.27 & 0.16 & 23.8 & 80.2 & 29.74 \\
 & InstructIR & 20.37 & 0.62 & 0.51 & 0.27 & 0.19 & 0.27 & 26.93 & 0.32 & 0.27 & 13.2 & \textcolor{blue}{90.4} & 22.54 \\
 & AutoDIR & 21.66 & 0.72 & 0.32 & 0.20 & 0.10 & 0.21 & 13.30 & 0.22 & 0.13 & 42.5 & 76.1 & 45.03 \\
 \cmidrule(l){2-14}
 & AgenticIR & 21.72 & 0.72 & 0.30 & 0.19 & 0.11 & 0.16 & 9.28 & 0.16 & 0.05 & 53.4 & 47.2 & 46.96 \\
 & MAIR & 20.92 & 0.70 & 0.28 & - & 0.29 & 0.14 & 7.87 & - & - & - & - & - \\
 & 4KAgent & 20.95 & 0.67 & 0.30 & 0.18 & 0.07 & 0.10 & 6.16 & 0.11 & 0.09 & 51.9 & 45.2 & 44.09 \\
 & TIR-Agent & \textcolor{blue}{22.80} & \textcolor{blue}{0.73} & 0.27 & - & 0.24 & 0.10 & 3.61 & - & - & - & - & - \\
 & SEAR & 22.13 & \textcolor{blue}{0.73} & 0.29 & - & 0.30 & 0.14 & 8.20 & - & - & - & - & - \\
 & OPERA & 22.28 & \textcolor{red}{0.75} & 0.31 & - & 0.31 & 0.19 & 9.75 & - & - & - & - & - \\
 \cmidrule(l){2-14}
 & GPT-Image-2.5 & 18.42 & 0.56 & 0.34 & 0.19 & \textcolor{blue}{0.02} & 0.11 & 4.65 & 0.07 & 0.02 & 45.4 & 22.1 & 30.42 \\
 & HarnessIR-GPT & \textbf{19.96} & \textbf{0.60} & \textbf{0.28} & \textbf{0.16} & \textcolor{blue}{\textbf{0.02}} & \textcolor{blue}{\textbf{0.08}} & \textbf{3.36} & \textcolor{red}{\textbf{0.03}} & \textcolor{blue}{\textbf{0.01}} & \textbf{69.6} & \textbf{40.2} & \textbf{51.27} \\
 \cmidrule(l){2-14}
 & Nano Banana 2 & 21.23 & 0.66 & \textcolor{blue}{0.22} & \textcolor{blue}{0.12} & \textcolor{red}{0.00} & \textcolor{red}{0.03} & \textcolor{blue}{0.67} & \textcolor{blue}{0.04} & \textcolor{blue}{0.01} & \textcolor{red}{79.9} & 50.2 & \textcolor{blue}{61.00} \\
 & HarnessIR-NB2 & \textcolor{red}{\textbf{22.94}} & \textbf{0.68} & \textcolor{red}{\textbf{0.19}} & \textcolor{red}{\textbf{0.11}} & \textcolor{red}{\textbf{0.00}} & \textcolor{red}{\textbf{0.03}} & \textcolor{red}{\textbf{0.14}} & \textcolor{blue}{\textbf{0.04}} & \textcolor{red}{\textbf{0.00}} & \textcolor{blue}{78.1} & \textbf{58.2} & \textcolor{red}{\textbf{65.22}} \\
 \cmidrule(l){2-14}
 & GT & - & - & - & - & 0.64 & 0.65 & 68.85 & 0.64 & -0.92 & - & - & - \\
& LQ & 20.64 & 0.60 & 0.54 & 0.29 & 0.19 & 0.26 & 27.62 & 0.33 & 0.25 \\
\midrule
\multirow{18}{*}{Group C}
 & AirNet & 20.22 & 0.59 & 0.57 & 0.30 & 0.30 & 0.35 & 39.18 & 0.41 & 0.35 & 26.0 & \textcolor{red}{97.6} & 33.00 \\
 & PromptIR & 20.30 & 0.59 & 0.57 & 0.30 & 0.30 & 0.35 & 39.39 & 0.42 & 0.35 & 26.0 & \textcolor{blue}{94.1} & 31.15 \\
 & MiOIR & 19.36 & 0.58 & 0.56 & 0.30 & 0.30 & 0.37 & 37.29 & 0.43 & 0.34 & 18.9 & 92.2 & 27.25 \\
 & DA-CLIP & 19.35 & 0.57 & 0.58 & 0.31 & 0.31 & 0.36 & 39.27 & 0.43 & 0.34 & 14.1 & 89.9 & 20.81 \\
 & InstructIR & 20.63 & 0.55 & 0.64 & 0.32 & 0.31 & 0.35 & 40.73 & 0.44 & 0.41 & 28.2 & 93.4 & 33.76 \\
 & AutoDIR & 19.06 & 0.57 & 0.47 & 0.27 & 0.19 & 0.30 & 22.29 & 0.30 & 0.19 & 34.2 & 61.6 & 34.00 \\
 \cmidrule(l){2-14}
 & AgenticIR & 20.20 & 0.59 & 0.43 & 0.25 & 0.20 & 0.20 & 17.10 & 0.24 & 0.10 & 39.9 & 35.6 & 33.87 \\
 & MAIR & 19.42 & 0.55 & 0.41 & - & 0.36 & 0.23 & 17.49 & - & - & - & - & - \\
 & 4KAgent & 19.77 & 0.56 & 0.43 & 0.24 & 0.14 & 0.13 & 13.29 & 0.15 & 0.16 & 40.1 & 31.1 & 29.78 \\
 & TIR-Agent & 19.83 & 0.56 & 0.41 & - & 0.27 & 0.10 & 11.91 & - & - & - & - & - \\
 & SEAR & 20.61 & 0.59 & 0.42 & - & 0.37 & 0.21 & 16.12 & - & - & - & - & - \\
 & OPERA & \textcolor{blue}{20.75} & \textcolor{red}{0.61} & 0.47 & - & 0.36 & 0.28 & 18.94 & - & - & - & - & - \\
 \cmidrule(l){2-14}
 & GPT-Image-2.5 & 17.96 & 0.52 & 0.36 & 0.20 & 0.02 & 0.09 & 4.48 & 0.05 & 0.02 & 58.2 & 20.5 & 33.20 \\
 & HarnessIR-GPT & \textbf{18.95} & \textbf{0.54} & \textbf{0.34} & \textbf{0.18} & \textcolor{red}{\textbf{0.00}} & \textcolor{blue}{\textbf{0.07}} & \textbf{2.21} & \textcolor{red}{\textbf{0.02}} & 0.03 & \textbf{61.8} & \textbf{27.4} & \textbf{39.25} \\
 \cmidrule(l){2-14}
 & Nano Banana 2 & 19.81 & 0.58 & \textcolor{blue}{0.28} & \textcolor{blue}{0.15} & \textcolor{red}{0.00} & \textcolor{red}{0.03} & \textcolor{blue}{1.07} & \textcolor{blue}{0.04} & \textcolor{blue}{0.01} & \textcolor{red}{70.0} & 33.0 & \textcolor{blue}{44.85} \\
 & HarnessIR-NB2 & \textcolor{red}{\textbf{21.23}} & \textcolor{blue}{\textbf{0.60}} & \textcolor{red}{\textbf{0.25}} & \textcolor{red}{\textbf{0.14}} & \textcolor{blue}{0.01} & \textcolor{red}{\textbf{0.03}} & \textcolor{red}{\textbf{0.29}} & \textcolor{blue}{\textbf{0.04}} & \textcolor{red}{\textbf{0.00}} & \textcolor{blue}{68.8} & \textbf{38.4} & \textcolor{red}{\textbf{49.12}} \\
 \cmidrule(l){2-14}
 & GT & - & - & - & - & 0.64 & 0.65 & 68.85 & 0.64 & -0.92 & - & - & - \\
 & LQ & 18.73 & 0.50 & 0.69 & 0.35 & 0.29 & 0.31 & 41.64 & 0.44 & 0.41 \\
\bottomrule
\end{tabular}}
\end{table}

Tab.~\ref{tab:mio100_abc} shows the detailed quantitative results for each group of MiO100. Restoration becomes more challenging as the simulated degradations become more complex. Each of Groups A and B combines two degradations, while Group C combines three, resulting in more severe and interacting distortions. Accordingly, most methods show lower PSNR, SSIM, and DF-Score for Group C. For example, when using Nano Banana 2 directly, the DF-Score falls from $61.75$ in Group A and $61.00$ in Group B to $44.85$ in Group C.

HarnessIR improves both executors in all three groups. For NB2, it raises the DF-Score to $65.47$, $65.22$, and $49.12$ for Groups A, B, and C, respectively. The gains over direct GPT-Image-2.5 are larger for Groups A and B, at $27.29$ and $20.85$ points, and smaller for Group C, at $6.05$ points. For all three groups, both HarnessIR variants achieve higher DF-Scores than the previous methods. Overall, the results show that HarnessIR's advantage persists across different degradation levels.

\section{More Visual Results}
\label{app:visual}

\begin{figure*}[t]
    \centering
        \includegraphics[width=0.93\linewidth]{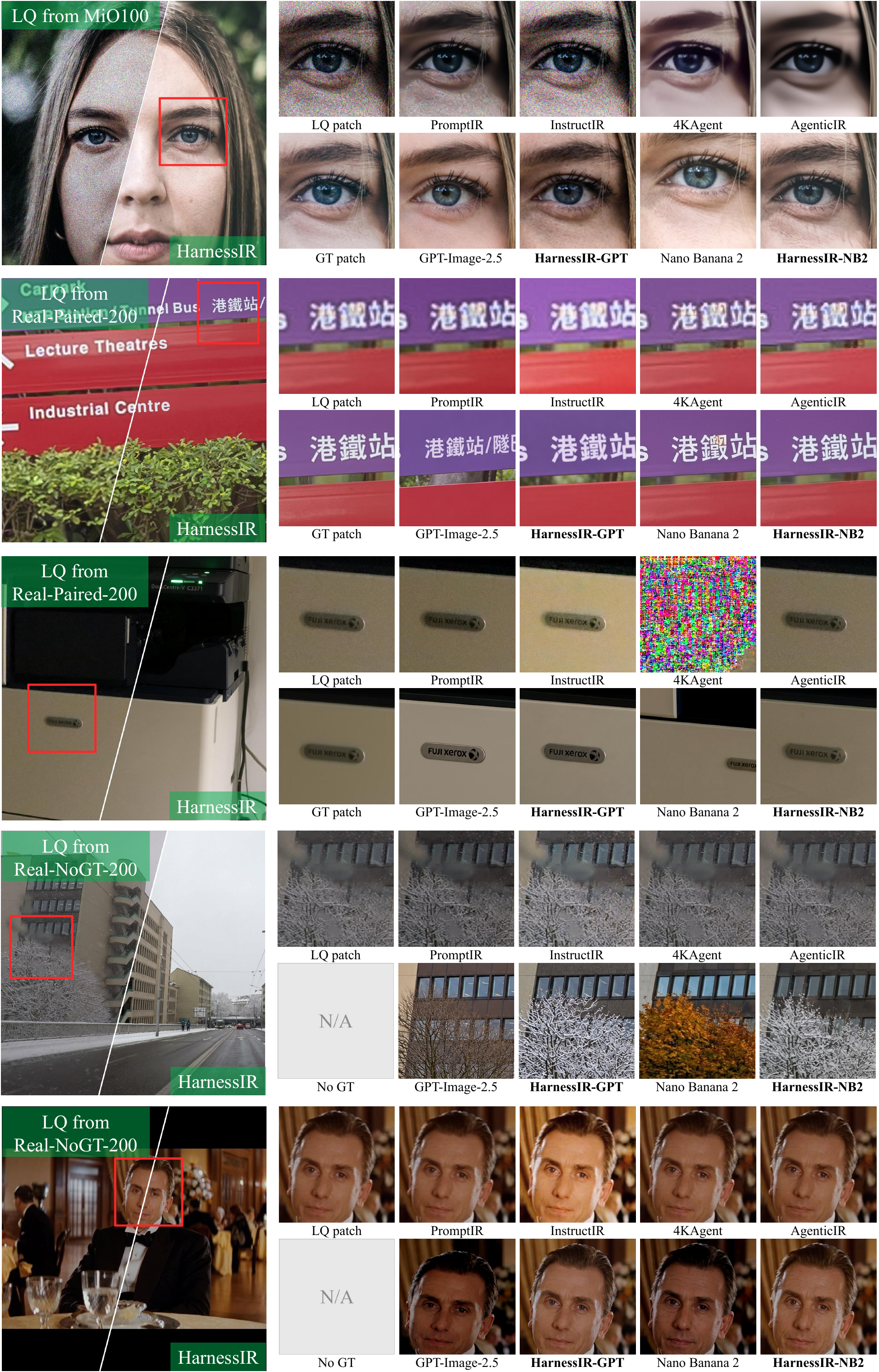}
        \vspace{-3mm}
        \caption{Additional visual comparisons of different methods. HarnessIR more effectively removes degradations while preserving image content. Zoom in for a better view.}
        \label{fig:morevisual}
    \vspace{-3mm}
\end{figure*}

To further demonstrate the visual quality of HarnessIR, Fig.~\ref{fig:morevisual} presents additional comparisons on the three test sets. Compared with direct MFM usage, HarnessIR restores facial details more faithfully and maintains a consistent color appearance, as illustrated by the first and fifth examples. It also recovers text while avoiding the spatial shifts and content changes observed in the direct MFM results, particularly in the second and third examples. For rainy and snowy scenes, HarnessIR removes the rain or snow that obstructs visibility while preserving the structure and appearance of other scene elements. For instance, it does not alter the shape of trees or hallucinate leaves in the fourth example of Fig.~\ref{fig:morevisual}.

The comparisons also reveal several limitations of previous methods on real-world images. Existing all-in-one and agentic methods often provide only a limited visible improvement on such inputs. InstructIR frequently brightens scenes even when low-light enhancement is not requested, as shown in the third and fifth examples. 4KAgent also suffers from a severe failure in the third example, producing a substantially corrupted result. This behavior illustrates a limitation of sequentially applying single-degradation restoration models: processing one degradation can change the distribution of the remaining degradations, causing the next model in the sequence to encounter an out-of-distribution input and fail. HarnessIR avoids this error-propagation problem by handling mixed degradations in a unified MFM execution while applying a redo refinement strategy.

\end{document}